\documentclass[journal]{IEEEtran}
\usepackage{times}
\usepackage{epsfig}
\usepackage{graphicx}
\usepackage{amsmath}
\usepackage{amssymb}
\usepackage{mathtools}
\usepackage[caption=false,font=footnotesize]{subfig}
\usepackage{gensymb}
\usepackage{float}
\usepackage{booktabs}
\usepackage{multirow}
\usepackage[numbers,sort&compress]{natbib}
\usepackage{threeparttable}
\usepackage{algorithm}
\usepackage{algpseudocode}

\usepackage[pagebackref=true,breaklinks=true,letterpaper=true,colorlinks,bookmarks=false]{hyperref}
\hypersetup{
  colorlinks, linkcolor=red
}

\newcommand{\bx}{\mathbf{x}}

\newcommand{\bs}{\mathbf{s}}
\newcommand{\bS}{\mathbf{S}}

\newcommand{\IR}{\mathbb{R}}

\newcommand{\EL}{\mathcal{L}}
\DeclareRobustCommand{\eg} {\textit{e}.\textit{g}.}

\DeclareRobustCommand{\NAME}{\textit{SSR-FCN}~}
\DeclareRobustCommand{\NAMENOSPACE}{\textit{SSR-FCN}}
\DeclareRobustCommand{\etal}{\textit{et al.}}

\algblock{Input}{EndInput}
\algnotext{EndInput}
\algblock{Output}{EndOutput}
\algnotext{EndOutput}

\makeatletter
\newcounter{phase}[algorithm]
\newlength{\phaserulewidth}
\newcommand{\setphaserulewidth}{\setlength{\phaserulewidth}}

\makeatother

\setphaserulewidth{.7pt}

\usepackage{capt-of,etoolbox}

\begin{document}
%
\title{Look Locally Infer Globally:\\
A Generalizable Face Anti-Spoofing Approach}
%
%
%

\author{Debayan~Deb,~\IEEEmembership{Student Member,~IEEE,}
        and~Anil~K.~Jain,~\IEEEmembership{Life~Fellow,~IEEE}
\thanks{M. Shell was with the Department
of Electrical and Computer Engineering, Georgia Institute of Technology, Atlanta,
GA, 30332 USA e-mail: (see http://www.michaelshell.org/contact.html).}}

\twocolumn[{%
\renewcommand\twocolumn[1][]{#1}%
\begin{@twocolumnfalse}
    \maketitle
  \end{@twocolumnfalse}
\begin{center}
\footnotesize
    \centering
    \begin{minipage}{0.13\linewidth}
    \includegraphics[width=\linewidth]{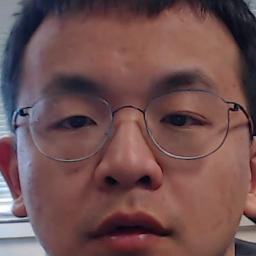}
    \centering {(a) Live}
    \end{minipage}
    \begin{minipage}{0.13\linewidth}
    \includegraphics[width=\linewidth]{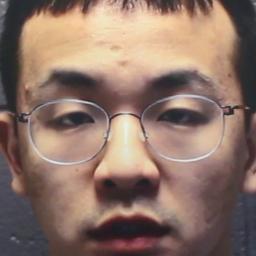}
    \centering {(b) Print}
    \end{minipage}
    \begin{minipage}{0.13\linewidth}
    \includegraphics[width=\linewidth]{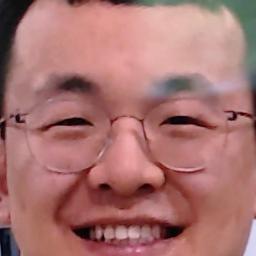}
    \centering {(c) Replay}
    \end{minipage}
    \begin{minipage}{0.13\linewidth}
    \includegraphics[width=\linewidth]{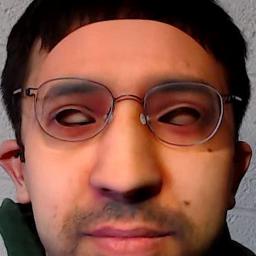}
    \centering {(d) Half Mask}
    \end{minipage}
    \begin{minipage}{0.13\linewidth}
    \includegraphics[width=\linewidth]{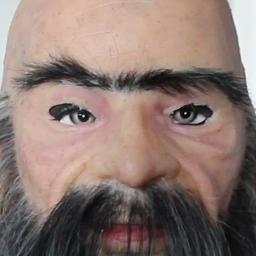}
    \centering {(e) Silicone Mask}
    \end{minipage}
    \begin{minipage}{0.13\linewidth}
    \includegraphics[width=\linewidth]{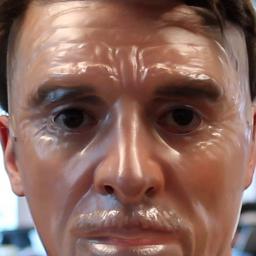}
    \centering {(f) Transparent}
    \end{minipage}
    \begin{minipage}{0.13\linewidth}
    \includegraphics[width=\linewidth]{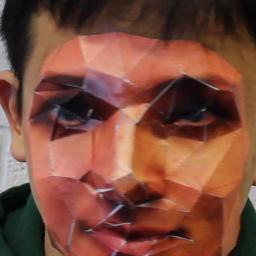}
    \centering {(g) Paper Mask}
    \end{minipage}\\ \vspace{0.3em}
    \begin{minipage}{0.13\linewidth}
    \includegraphics[width=\linewidth]{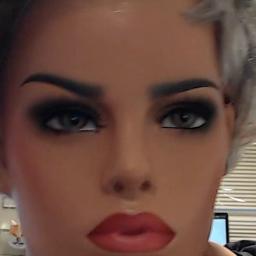}
    \centering {(h) Mannequin}
    \end{minipage}
    \begin{minipage}{0.13\linewidth}
    \includegraphics[width=\linewidth]{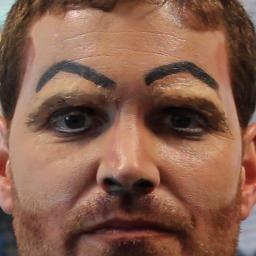}
    \centering {(i) Obfuscation}
    \end{minipage}
    \begin{minipage}{0.13\linewidth}
    \includegraphics[width=\linewidth]{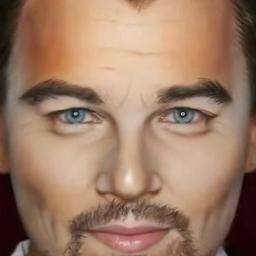}
    \centering {(j) Impersonation}
    \end{minipage}
    \begin{minipage}{0.13\linewidth}
    \includegraphics[width=\linewidth]{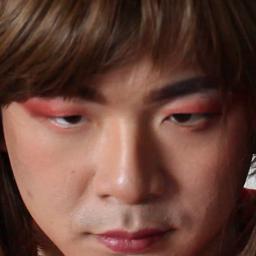}
    \centering {(k) Cosmetic}
    \end{minipage}
    \begin{minipage}{0.13\linewidth}
    \includegraphics[width=\linewidth]{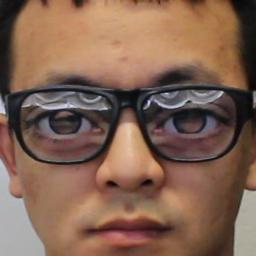}
    \centering {(l) FunnyEye}
    \end{minipage}
    \begin{minipage}{0.13\linewidth}
    \includegraphics[width=\linewidth]{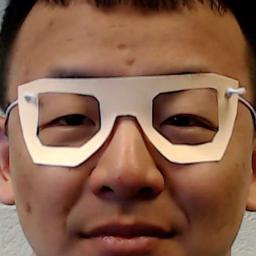}
    \centering {(m) Paper Glasses}
    \end{minipage}
    \begin{minipage}{0.13\linewidth}
    \includegraphics[width=\linewidth]{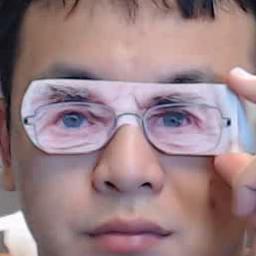}
    \label{fig:paper_cut}
    \centering {(n) Paper Cut}
    \end{minipage}\\
    \begin{minipage}{\linewidth}
    \begin{figure}[H]
    \captionof{figure}{Example spoof attacks: Simple attacks include (b) printed photograph, or (c) replaying the victim's video More advanced spoofs can also be leveraged such as (d-h) 3D masks, (i-k) make-up attacks, or (l-n) partial attacks~\cite{siw}. A live face is shown in (a) for comparison. Here, the spoofs in (b-c, k-n) belong to the same person in (a).}
    \label{fig:spoof_types}
    \end{figure}
    \end{minipage}
\end{center}%
}]

{
  \renewcommand{\thefootnote}%
    {\fnsymbol{footnote}}
  \footnotetext{D. Deb and A. K. Jain are with the Department of Computer Science and Engineering, Michigan State University, East Lansing, MI, 48824. E-mail: \{debdebay, jain\}@cse.msu.edu}
}

\begin{abstract}
State-of-the-art spoof detection methods tend to overfit to the spoof types seen during training and fail to generalize to unknown spoof types. Given that face anti-spoofing is inherently a local task, we propose a face anti-spoofing framework, namely Self-Supervised Regional Fully Convolutional Network (\NAMENOSPACE), that is trained to learn local discriminative cues from a face image in a self-supervised manner. The proposed framework improves generalizability while maintaining the computational efficiency of holistic face anti-spoofing approaches ($<$ 4 ms on a Nvidia GTX 1080Ti GPU). The proposed method is interpretable since it localizes which parts of the face are labeled as spoofs. Experimental results show that \NAME can achieve TDR = 65\% @ 2.0\% FDR when evaluated on a dataset comprising of 13 different spoof types under unknown attacks while achieving competitive performances under standard benchmark datasets (Oulu-NPU, CASIA-MFSD, and Replay-Attack).

\end{abstract}


\IEEEpeerreviewmaketitle

\section{Introduction}
\IEEEPARstart{T}{he} accuracy, usability, and touchless acquisition of state-of-the-art automated face recognition systems (AFR) have led to their ubiquitous adoption in a plethora of domains, including mobile phone unlock, access control systems, and payment services. The adoption of deep learning models over the past decade has led to prevailing AFR systems with accuracies as high as 99\% True Accept Rate at 0.1\% False Accept Rate~\cite{nist}. Despite this impressive recognition performance, current AFR systems remain vulnerable to the growing threat of \emph{spoofs}\footnote{ISO standard IEC 30107-1:2016(E) defines spoofs as \emph{``presentation to the biometric data capture subsystem with the goal of interfering with the operation of the biometric system"}~\cite{iso}. Note that these spoofs are different from digital manipulation of face images, such as DeepFakes~\cite{deep_fake} and adversarial faces~\cite{advfaces}.}.

Face spoofs are ``fake faces" which can be constructed with a variety of different materials,~\eg, 3D printed masks, printed paper, or digital devices (video replay attacks from a mobile phone) with a goal of enabling a hacker to impersonate a victim's identity, or alternatively, obfuscate their own identity (see Figure~\ref{fig:spoof_types}). With the rapid proliferation of face images/videos on the Internet (especially on social media websites, such as Facebook, Twitter, or LinkedIn), replaying videos containing the victim's face or presenting a printed photograph of the victim to the AFR system is a trivial task~\cite{handbook_spoof}. Even if a face spoof detection system could trivially detect printed photographs and replay video attacks (\eg, with depth sensors), attackers can still attempt to launch more sophisticated attacks such as 3D masks~\cite{silicone}, make-up, or even virtual reality~\cite{VR}. 


\begin{figure*}[!t]
    \centering
    \includegraphics[width=\textwidth]{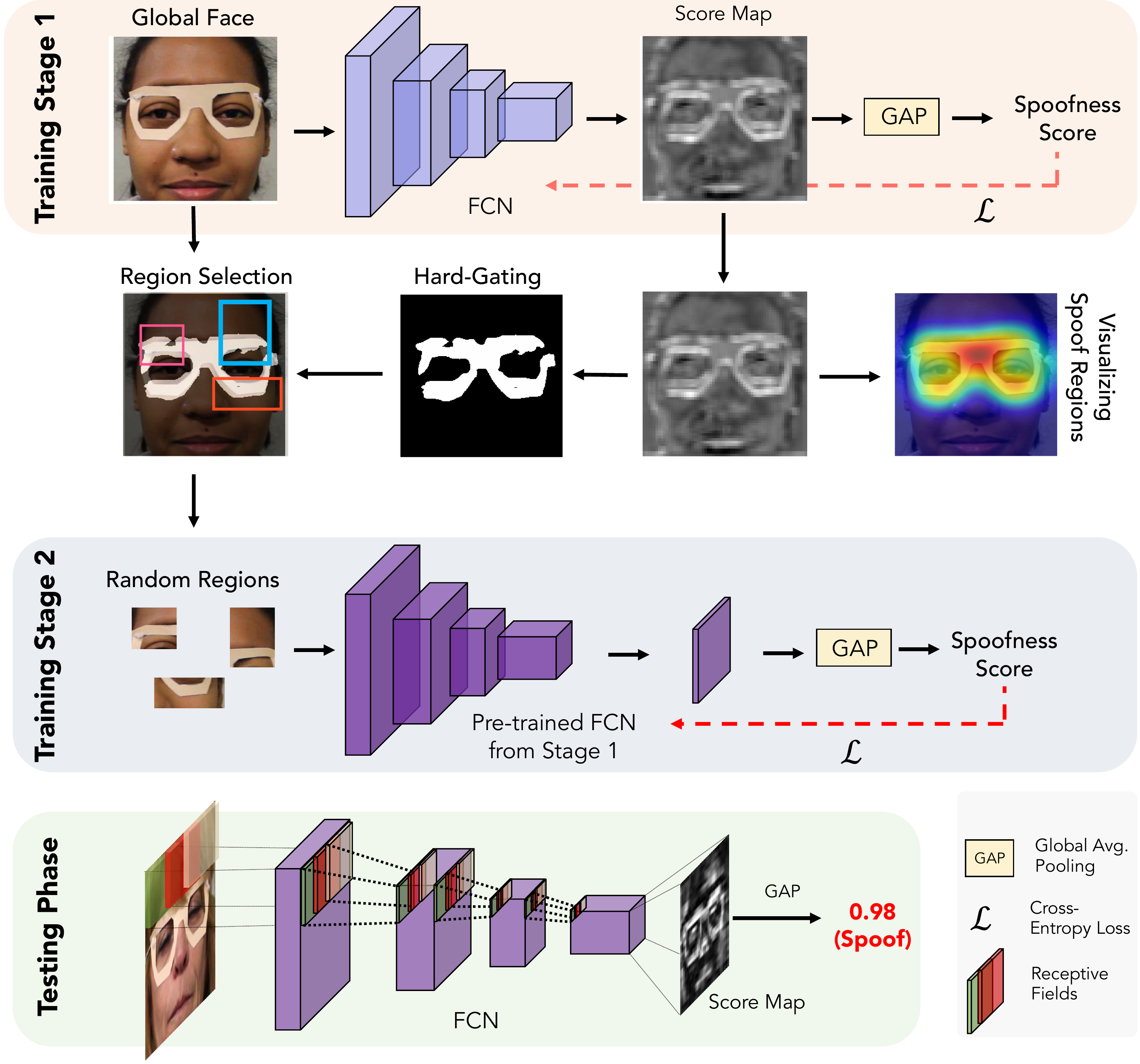}
    \caption{An overview of the proposed Self-Supervised Regional Fully Convolution Network (\NAMENOSPACE). We train in two stages: (1) Stage 1 learns global discriminative cues via training on the entire face image. The score map obtained from stage 1 is hard-gated to obtain spoof regions in the face image. We randomly crop arbitrary-size patches from the spoof regions and fine-tune our network in stage 2 to learn local discriminative cues. During test, we input the entire face image to obtain the classification score. The score map can also be used to visualize the spoof regions in the input image.}
    \label{fig:overview}
\end{figure*}

The need for preventing face spoofs is becoming increasingly urgent due to the user's privacy concerns associated with spoofed systems. Failure to detect face spoofs can be a major security threat due to the widespread adoption of automated face recognition systems for border control~\cite{fastpass}. In 2011, a young individual from Hong Kong boarded a flight to Canada disguised as an old man with a flat hat by wearing a silicone face mask to successfully fool the border control authorities~\cite{real_spoof}.
Also consider that, with the advent of Apple's iPhone X and Samsung's Galaxy S8, all of us are carrying automated face recognition systems in our pockets embedded in our smartphones. Face recognition on our phones facilitates (i) unlocking the device, (ii) conducting financial transactions, and (iii) access to privileged content stored on the device. Failure to detect spoof attacks on smartphones could compromise confidential information such as emails, banking records, social media content, and personal photos~\cite{iphonex_spoof}.

\begin{table*}[!t] 
\footnotesize
\caption{A summary of publicly available face anti-spoofing datasets.}
\centering
\resizebox{\textwidth}{!}{\begin{tabular}{l||c||c|c||c|c|c||c|c|c|c|c||c}
\noalign{\hrule height 1.5pt}
\textbf{Dataset} &\textbf{Year} &\multicolumn{2}{c||}{\textbf{Statistics}} &\multicolumn{3}{c||}{\textbf{PIE Variations}} &\multicolumn{5}{c||}{\textbf{\# Spoof Types}} & \textbf{Total}\\ \hline
& & \textbf{\# Subj.} & \textbf{\# Vids.} & \textbf{Pose} & \textbf{Expression Change} & \textbf{Illumination Change} & \textbf{Replay} & \textbf{Print} & \textbf{3D Mask} & \textbf{Makeup} & \textbf{Partial} & \\ \hline
 Replay-Attack~\cite{replay_attack} & 2012 & 50 & 1,200 & Frontal & No & Yes & 2 & 1 & 0 & 0 & 0 & 3\\
 CASIA-FASD~\cite{casia_mfsd}  & 2012 & 50 & 600 & Frontal  & No & No & 1 & 1 & 0 & 0 & 0 & 2\\
 3DMAD~\cite{3dmad}  & 2013 & 17 & 255 & Frontal & No & No & 0 & 0 & 1 & 0 & 0 & 1\\
 MSU-MFSD~\cite{msu_mfsd}  & 2015 & 35 & 440 & Frontal  & No & No & 2 & 1 & 0 & 0 & 0 & 3\\
 Replay-Mobile~\cite{replay-mobile}  & 2016 & 40 & 1,030 & Frontal & No & Yes & 1 & 1 & 0 & 0 & 0 & 2\\
 HKBU-MARs~\cite{hkbu} & 2016 & 35 & 1,009 & Frontal & No & Yes & 0 & 0 & 2 & 0 & 0 & 2\\
 Oulu-NPU~\cite{oulu_npu}  & 2017 & 55 & 4,950 & Frontal & No & Yes & 2 & 2 & 0 & 0 & 0 & 4\\
 SiW~\cite{siw}  & 2018 & 165 & 4,620 & $[-90\degree, 90\degree]$ & Yes & Yes & 4 & 2 & 0 & 0 & 0 & 6\\
 SiW-M~\cite{deep_tree} & 2019 & 493 & 1,630 & $[-90\degree, 90\degree]$ & Yes & Yes & 1 & 1 & 5 & 3 & 3 & 13\\
  \hline
\noalign{\hrule height 1.5pt}
\end{tabular}}
\label{tab:public_datasets}
\end{table*}

With numerous approaches proposed to detect face spoofs, current face anti-spoofing methods have following shortcomings:
\paragraph{Generalizabilty} Since the exact type of spoof attack may not be known beforehand, how to generalize well to ``unknown"\footnote{We make a distinction between \emph{unseen} and \emph{unknown} attack types. Unseen attacks are spoof types that are known to the developers whereby algorithms can be specifically tailored to detect them, but their data is never used for training. Unknown attacks are spoof types that are not known to the developers and neither seen during training. In this paper, we focus on the more challenging scenario of unknown attacks.} attacks is of utmost importance. A majority of the prevailing state-of-the-art face anti-spoofing techniques focus only on detecting 2D printed paper and video replay attacks, and are vulnerable to spoofs crafted from materials not seen during training of the dectector. In fact, studies show a two-fold increase in error when spoof detectors encounter unknown spoof types~\cite{siw}. In addition, current face anti-spoofing approaches rely on densely connected neural networks with a large number of learnable parameters (exceeding $2.7M$), where the lack of generalization across unknown spoof types is even more pronounced. 
\paragraph{Lack of Interpretability} Given a face image, face anti-spoofing approaches typically output a holistic face~\emph{``spoofness score"} which depicts the likelihood that the input image is live or spoof. Without an ability to visualize which regions of the face contribute to the overall decision made be the network, the global spoofness score alone is not sufficient for a human operator to interpret the network's decision.


In an effort to impart generalizability and interpretability to face spoof detection systems, we propose a face anti-spoofing framework specifically designed to detect unknown spoof types, namely, \textbf{Self-Supervised Regional Fully Convolutional Network} (\NAMENOSPACE). A Fully Convolutional Network (FCN) is first trained to learn global discriminative cues and automatically identify spoof regions in face images. The network is then fine-tuned to learn local representations via regional supervision. Once trained, the deployed model can automatically locate regions where spoofness occurs in the input image and provide a final spoofness score. 

Our contributions are as follows:

\begin{itemize}
    \item We show that features learned from local face regions have better generalization ability than those learned from the entire face image alone.
    \item We provide extensive experiments to show that the proposed approach, \NAMENOSPACE, outperforms other local region extraction strategies and state-of-the-art face anti-spoofing methods on one of the largest publicly available dataset, namely, SiW-M, comprised of 13 different spoof types. The proposed method reduces the Equal Error Rate (EER) by (i) 14\% relative to state-of-the-art~\cite{cdc} under the unknown attack setting, and (ii) 40\% on known spoofs. In addition, \NAME achieves competitive performance on standard benchmarks on Oulu-NPU~\cite{oulu_npu} dataset and outperforms prevailing methods on cross-dataset generalization (CASIA-FASD~\cite{casia_mfsd} and Replay-Attack~\cite{replay_attack}).
    \item The proposed \NAME is also shown to be more interpretable since it can directly predict which parts of the faces are considered as spoofs. 
\end{itemize}

\section{Background}
In order to mitigate the threats associated with spoof attacks, numerous face anti-spoofing techniques, based on both software and hardware solutions, have been proposed. Early software-based solutions utilized liveness cues, such as eye blinking, lip movement, and head motion, to detect print attacks~\cite{active1, active2, active3, active4}. However, these approaches fail when they encounter unknown attacks such as printed attacks with cut eye regions (see Figure~\ref{fig:spoof_types}n). In addition, these methods require active cooperation of user in providing specific types of images making them tedious to use.

\begin{figure*}[!t]
    \centering
    \includegraphics[width=\linewidth]{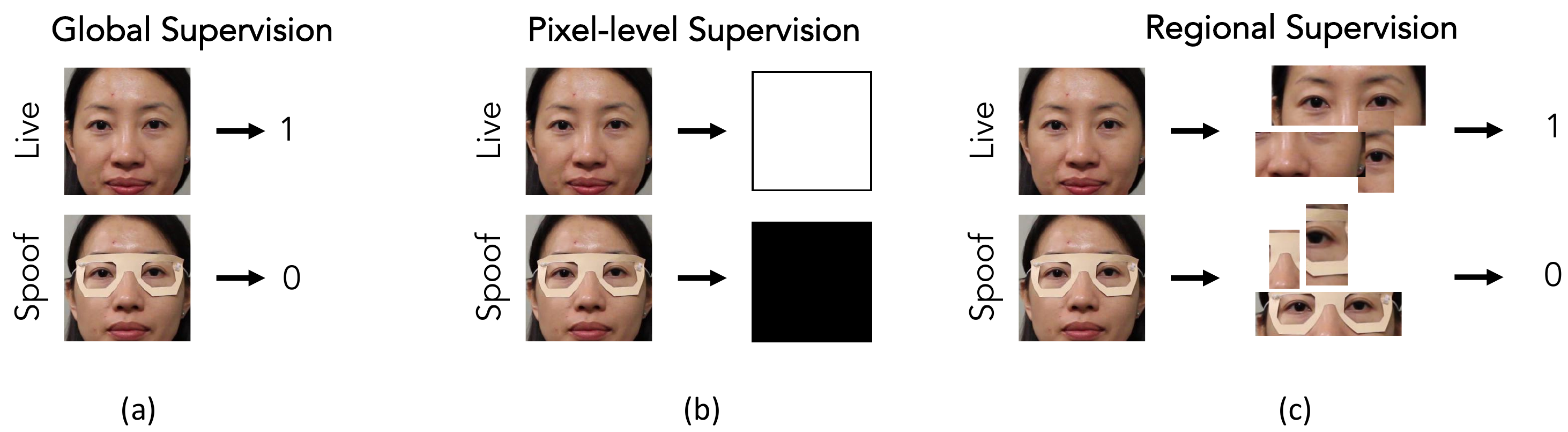}
    \caption{Illustration of drawbacks of prior approaches. (a) Top: example of a live face; Bottom: example of a paper glasses spoof. In this case, the spoof artifact is only present in the eye-region of the face. (b) Classifier trained with global supervision overfits to the live class since both images are mostly live  (the spoof instrument covers only a part of the face). (c) Pixel-level supervision assumes the entire image is spoof or live and constructs label maps accordingly. This is not a valid assumption in mask, makeup, and partial spoof types. Instead, (d) the proposed framework trains of extracted regions from face images. These regions can be based on domain knowledge, such as eye, nose, mouth regions, or randomly cropped.}
    \label{fig:drawbacks}
\end{figure*}

Since then, researchers have moved on to \emph{passive} face anti-spoofing approaches that rely on texture analysis for distinguishing lives and spoofs, rather than motion or liveness cues. The majority of face anti-spoofing methods only focus on detecting print and replay attacks, which can be detected using features such as color and texture~\cite{hog1, lbp1, learned_color, sift1, iqa1, iqa2}. Many prior studies employ handcrafted features such as 2D Fourier Spectrum~\cite{msu_mfsd, fourier}, Local Binary Patterns (LBP)~\cite{lbp1, lbp2, lbp3, lbp4}, Histogram of oriented gradients (HOG)~\cite{hog1, hog2}, Difference-of-Gaussians (DoG)~\cite{dog}, Scale-invariant feature transform (SIFT)~\cite{sift1}, and Speeded up robust features (SURF)~\cite{surf1}. Some techniques utilize anti-spoofing beyond the RGB color spectrum, such as incorporating the luminance and chrominance channels~\cite{lbp4}. Instead of a predetermined color spectrum, Li~\etal~\cite{learned_color} automatically learn a new color scheme that can best distinguish lives and spoofs. Another line of work extracts image quality features to detect spoof faces~\cite{iqa1, iqa2, msu_mfsd}. Due to the assumption that spoof types are one of replay or print attacks, these methods severely suffer from generalization to unknown spoof types.

Hardware-based solutions in literature have incorporated 3D depth information~\cite{sensor1, sensor2, casia_surf}, multi-spectral and infrared sensors~\cite{sensor3}, and even physiological sensors such as vein-flow information~\cite{sensor4}. Presentation attack detection can be further enhanced by incorporating background audio signals~\cite{sensor5}. However, with the inclusion of additional sensors along with a standard camera, the deployment costs can be exorbitant (\eg, thermal sensors for iPhones cost over USD $400$\footnote{\url{https://amzn.to/2zJ6YW4}}).

State-of-the-art face anti-spoofing systems utilize Convolutional Neural Networks (CNN) so the feature set (representation) is learned that best differentiates live faces from spoofs. Yang~\etal~were among the first to propose CNNs for face spoof detection and they showed about 70\% decrease in Half Total Error Rate (HTER) compared to the baselines comprised of handcrafted features~\cite{yang}. Further improvement in performance was achieved by directly modifying the network architecture~\cite{cnn1, cnn2, cnn3, cnn4}. Deep learning approaches also perform well for mask attack detection~\cite{silicone}. Incorporating auxiliary information (\eg~eye blinking) in deep networks can further improve the face spoof detection performance~\cite{active3, siw}.

 Table~\ref{tab:public_datasets} outlines the publicly available face anti-spoofing datasets.

\section{Motivation}
Our approach is motivated by following observations:
\subsubsection{Face Anti-Spoofing is a Local Task}
It is now generally accepted that for print and replay attacks, \emph{``face spoofing is usually a local task in which discriminative clues are ubiquitous and repetitive"}~\cite{despoof}. However, in the case of masks, makeups, and partial attacks, the ubiquity and repetitiveness of spoof cues may not hold true. For instance, in Figure~\ref{fig:drawbacks} (a-c), spoofing artifact (the paper glasses) are only present in the eye regions of the face. Unlike face recognition, face anti-spoofing does not require the entire face image in order to predict whether the image is a spoof or live. In fact, our experimental results and their analysis will confirm that the entire face image alone can adverserly affect the convergence and generalization of networks.

\subsubsection{Global vs. Local Supervision}
Prior work can be partitioned into two groups: (i) \emph{global supervision} where the input to the network is the entire face image and the CNN outputs a score indicating whether the image is live or spoof~\cite{yang, cnn1, cnn2, cnn3, cnn4, cnn5, patch, siw, cdc}, and (ii) \emph{pixel-level supervision} where multiple classification losses are aggregated over each pixel in the feature map~\cite{pixel, sun2020face}. These studies assume that all pixels in the face image is either live or spoof (see Figure~\ref{fig:drawbacks}(b)). This assumption holds true for spoof types, such as replay and print attacks (which are the only spoof types considered by the studies), but not for mask, makeup, and partial attacks. Therefore, pixel-level supervision can not only suffer from poor generalization across a diverse range of spoof types, but also convergence of the network is severely affected due to noisy labels.

In summary, based on the 13 different spoof types shown in Figure~\ref{fig:spoof_types}, for which we have the data, we gain the following insights: (i) face anti-spoofing is inherently a local task, and (ii) learning local representations can improve face anti-spoofing performance~\cite{pixel, sun2020face}. Motivated by (i), we hypothesize that utilizing a Fully Convolutional Network (FCN) may be more appropriate for the face anti-spoofing task compared to a traditional CNN. The second insight suggests FCNs can be intrinsically regularized to learn local cues by enforcing the network to~\emph{look} at local spatial regions of the face. In order to ensure that these regions mostly comprise spoof patterns, we propose a~\emph{self-supervised} region extractor.

\begin{table}[!t] 
\footnotesize
\caption{Architecture details of the proposed FCN backbone.}
\centering
\begin{tabular}{c|c|c|}
\noalign{\hrule height 1.5pt}
Layer & \# of Activations & \# of Parameters\\
\noalign{\hrule height 1.2pt}
Input & $H\times W\times3$ & 0\\  \hline
Conv & $H/2\times W/2\times64$ & $3\times3\times3\times64 + 64$\\ \hline
Conv & $H/4\times W/4\times128$ & $3\times3\times64\times128 + 128$\\ \hline
Conv & $H/8\times W/8\times256$ & $3\times3\times128\times256 + 256$\\ \hline
 Conv & $H/16\times H/16\times512$ & $3\times3\times256\times512 + 512$\\ \hline
Conv & $H/16\times H/16\times1$ & $3\times3\times512\times1 + 1$\\ \hline
GAP & 1 & 0\\
\noalign{\hrule height 1.1pt}
Total & & \textbf{1.5M}\\
\noalign{\hrule height 1.5pt}
\end{tabular}
\begin{tablenotes}
\item {\scriptsize \vspace{0.3em} Conv and GAP refer to convolutional and global average pooling operations.}
\end{tablenotes}
\label{tab:params}
\end{table}

\section{Proposed Approach}
In  this  section,  we  describe  the  proposed~\emph{Self-Supervised Regional Fully Convolutional Network} (\NAMENOSPACE)  for  generalized face anti-spoofing.  As shown in Figure~\ref{fig:overview}, we train the network in two stages, (a) Stage I learns global discriminative cues and predicts score maps, and (b) Stage II extracts arbitrary-size regions from spoof areas and fine-tunes the network via regional supervision.

\subsection{Network Architecture}
\label{sec:arch}
In typical image classification tasks, networks are designed such that information present in the input image can be used for learning~\emph{global} discriminative features in the form of a feature vector without utilizing the  spatial arrangement in the input. To this end, a fully connected (FC) layer is generally introduced at the end of the last convolutional layer. The FC layer is responsible for stripping away all spatial information and reducing the feature maps into a single D-dimensional feature vector. Given the plethora of available spoof types, it is better to learn~\emph{local} representations and make decisions on local spatial inputs rather than global descriptors. Therefore, we employ a Fully Convolutional Network (FCN) by replacing the FC layer in a traditional CNN with a $1\times1$ convolutional layer. This leads to three major advantages over traditional CNNs:
\begin{itemize}
    \item \textbf{Arbitrary-sized inputs}: The proposed FCN can accept input images of any image size. This property can be exploited to learn discriminative features at local spatial regions, regardless of the input size, rather than overfitting to a global representation of the entire face image.
    \item \textbf{Interpretability}: Since the proposed FCN is trained to provide decisions at a local level, the score map output by the network can be used to identify the spoof regions in the face.
    \item \textbf{Efficiency}: Via FCN, an entire face image can be inferred only once where local decisions are dynamically aggregated via the $1\times1$ convolution operator.
\end{itemize}

\subsection{Network Efficiency}

A majority of prior work on CNN-based face anti-spoofing employs architectures that are densely connected with thirteen convolutional layers~\cite{patch, siw, pixel, domain_adaptation, cdc}. Even with the placement of skip connections, the number of learnable parameters exceed $2.7M$. As we see in Table~\ref{tab:public_datasets}, only a limited amount of training data\footnote{The lack of large-scale publicly available face anti-spoofing datasets is due to the time and effort required along with privacy concerns associated in acquiring such datasets.} is generally available in face anti-spoofing datasets. Limited data coupled with the large number of trainable parameters causes current approaches to overfit, leading to poor generalization performance under unknown attack scenarios. Instead, we employ a shallower neural network comprising of only five convolutional layers with approximately $1.5M$ learnable parameters (see Table~\ref{tab:params}).

\begin{figure*}[!t]
    \centering
    \includegraphics[width=\linewidth]{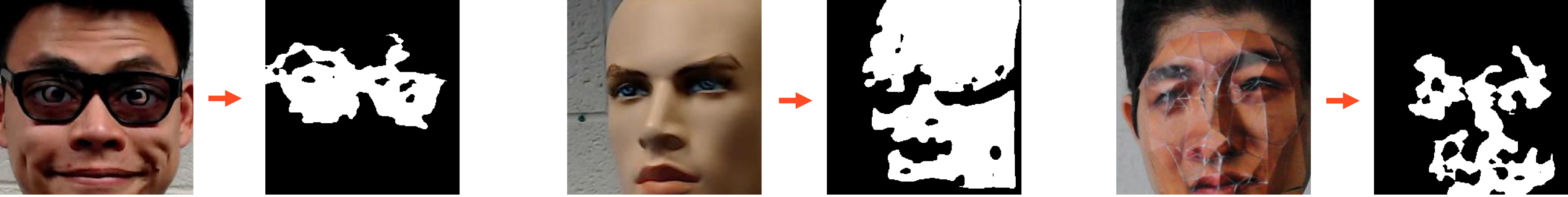}
    \caption{Three spoof images and their corresponding binary masks extracted from predicted score maps. Black regions correspond to predicted live regions, whereas, white regions indicate spoofness.}
    \label{fig:binary_masks}
\end{figure*}

\subsection{Stage I: Training FCN Globally}
\label{sec:global}
We first train the FCN with global face images in order to learn global discriminative cues and identify spoof regions in the face image. Given  an  image, $\bx\in\IR^{H\times W\times C}$,  we  detect  a  face  and  crop  the face region via 5 landmarks (two eyes, nose, and two mouth keypoints)  in  order  to  remove  background  information  not pertinent  to  the  task  of  face  anti-spoofing. Here, $H$, $W$, and $C$ refer to the height, width, and number of channels (3 in the case of RGB) of the input image. The  face  regions are then aligned and resized to a fixed-size (\eg, $256\times256$) in  order  to  maintain  consistent  spatial  information  across all training data. 

The proposed FCN consists of four downsampling convolutional blocks each coupled with batch normalization and ReLU activation. The feature map from the fourth convolutional layer passes through a $1\times1$ convolutional layer. The output of the $1\times1$ convolutional layer represents a score map $\bS \in \IR^{(H_S\times W_S\times1)}$ where each pixel in $\bS$ represents a live/spoof decision corresponding to its receptive field in the image. The height ($H_S$) and width ($W_S$) of the score map is determined by the input image size and the number of downsampling layers. For a $256\times256\times3$ input image, our proposed architecture outputs a $16\times16\times1$ score map.

The score map is then reduced to a single scalar value by globally average pooling. That is, the final classification score ($\bs$) for an input image is obtained from the $(H_S\times W_S\times1)$ score map ($\bS)$ by,
\begin{align}
\bs = \frac{1}{H_S\times W_S}\sum_{i}^{H_S}\sum_{j}^{W_S}\bS_{i,j}
\label{eqn:classification_score}
\end{align}
Using sigmoid activation on the final classification output ($\bs$), we obtain a scalar $p(c | \bx)\in [0, 1]$ predicting the likelihood that the input image is a spoof, where $c=0$ indicates live and $c=1$ indicates spoof. 

We train the network by minimizing the Binary Cross Entropy (BCE) loss,
\begin{align}
    \EL = -\left[y\times log(p(c | \bx)) + (1-y)\times log(1-p(c | \bx))\right]
    \label{eqn:bce}
\end{align}
where $y$ is the ground truth label of the input image. 

\subsection{Stage II: Training FCN on Self-Supervised Regions}
In order to supervise training at a local level, we propose a regional supervision strategy. We train the network to learn local cues by only showing certain regions of the face where spoof patterns exists. In order to ensure that spoof artifacts/patterns indeed exists within the selected regions, the pre-trained FCN from Stage I (\ref{sec:global}) can automatically guide the region selection process in spoof images. For live faces, we can randomly crop a region from any part of the image.

Due to the absence of a fully connected layer, notice that FCN naturally encodes decisions at each pixel in feature map $\bS$. In other words, higher intensity pixels within $\bS$ indicate a larger likelihood of a spoof pattern residing within the receptive field in the image. Therefore, discriminative regions (spoof areas) are automatically highlighted in the score map by training on entire face images (see Figure~\ref{fig:overview}). 

We can then craft a binary mask $M$ indicating the live/spoof regions in the input spoof images.  First, we first soft-gate the score map by min-max normalization such that we can obtain a score map in $[0,1]$, 
\begin{align}
    \bS' = \frac{(\bS - \min(\bS))}{(\max(\bS) - \min(\bS))}
     \label{eqn:normalize}
\end{align}
Let $f_{\bS'}(i,j)$ represent the activation in the $(i,j)$-th  spatial location in the scaled score map $\bS'$. The binary mask $M$ is designed by hard-gating,
\begin{align}
 \label{eqn:binary_mask}
    M(i,j) &=
    \begin{cases}
        1, & \text{if } \bS'(i,j) \geq \tau\\
        0, & \text{otherwise}\\
     \end{cases}
\end{align}
     
where, $\tau$ the size of the hard-gated region ($\tau=0.5$ in our case). A larger $\tau$ will result in small regions and smaller $\tau$ can lead to spurious spoof regions. Examples of binary masks are shown in Figure~\ref{fig:binary_masks}. From the binary mask, we can then randomly extract a rectangular bounding box such that the center of the rectangle lies within the detected spoof regions. In this manner, we can crop rectangular regions of arbitrary sizes from the input image such that each region contains spoof artifacts according to our pre-trained global FCN. We constrain the width and height of the bounding boxes to be between $MIN_{region}$ and $MAX_{region}$.

In this manner, we fine-tune our network to learn local discriminative cues. 


\subsection{Testing}
Since FCNs can accept abitrary input sizes and the fact that the proposed FCN has encountered entire faces in Stage I, we input the global face into the trained network and obtain the score map. The score map is then average pooled to extract the final classification output, which is then normalized by a sigmoid function in order to obtain a spoofness score within $[0,1]$. That is, the final classification score is obtained by,
\begin{align*}
    \frac{1}{1 + \exp(-\bs)}
\end{align*}

In addition to the classification score, the score map ($\bS$) can also be utilized for visualizing the spoof regions in the face by constructing a heatmap (see Figure~\ref{fig:overview}).

\section{Experiments}

\subsection{Datasets}
The following four datasets are utilized in our study (Table~\ref{tab:public_datasets}):
\subsubsection{Spoof-in-the-Wild with Multiple Attacks (SiW-M)~\cite{deep_tree}} A dataset, collected in 2019, comprising 13 different types of spoofs, acquired specifically for evaluating generalization performance on unknown spoof types. Compared with other publicly available datasets (Table~\ref{tab:public_datasets}), SiW-M is diverse in spoof attack types, environmental conditions, and face poses. We evaluate \NAME under both \emph{unknown} and \emph{known} settings, and perform ablation studies on this dataset.
\subsubsection{Oulu-NPU~\cite{oulu_npu}} A dataset comprised of 4,950 high-resolution video clips of 55 subjects. Oulu-NPU defines four protocols each designed for evaluating generalization against variations in capturing conditions, attack devices, capturing devices and their combinations. We use this dataset for comparing our approach with the prevailing state-of-the-art face anti-spoof methods on the four protocols.
\subsubsection{CASIA-FASD~\cite{casia_mfsd} \& Replay-Attack~\cite{replay_attack}} Both datasets, collected in 2012, are frequently employed in face anti-spoofing literature for testing \emph{cross-dataset generalization} performance. These two datasets provide a comprehensive collection of attacks, including warped photo attacks, cut photo attacks, and video replay attacks. Low-quality, normal-quality,and high-quality videos are recorded under different lighting conditions.

emph{All images shown in this paper are from SiW-M testing sets.}



\subsection{Data Preprocessing} For all datasets, we first extract all frames in a video. The frames are then passed  through MTCNN face detector~\cite{mtcnn} to detect 5 facial landmarks (two eyes, nose and two mouth corners).  Similarity transformation is used to align the face images based on the five landmarks. After transformation, the images are cropped to $256\times 256$. \emph{All face images shown in the paper are cropped and aligned.} Before passing into network, we normalize the images by requiring each pixel to be within $[-1, 1]$ by subtracting $127.5$ and dividing by $128.0$.

\subsection{Implementation Details} \NAME is implemented in Tensorflow, and trained with a constant learning rate of $(1e-3)$ with a mini-batch size of $128$. The objective function, $\EL$, is minimized using Adam optimizer~\cite{adam}. It takes 20 epochs to converge. Following~\cite{siw}, we randomly initialize all the weights of the convolutional layers using a normal distribution of $0$ mean and $0.02$ standard deviation. We restrict the self-supervised regions to be at least $1/4$ of the entire image, that is, $MIN_{region} = 64$ and at most $MAX_{region}=256$ which is the size of the global face image. Data augmentation during training involves random horizontal flips with a probability of $0.5$. For evaluation, we compute the spoofness scores for all frames in a video and temporally average them to obtain the final classification score.
\emph{For all experiments, we use a threshold of $0.5$ as our live/spoof decision threshold.}

\subsection{Evaluation Metrics} There is no standard metric used in literature for evaluating face spoof detection and each dataset provides their own evaluation protocol. For a fair comparison with prior work, we report the Average Classification Error Rate (ACER) (standardized in ISO/IEC 30107~\cite{iso}), Equal Error Rate (EER), and True Detection Rate (TDR) at 2.0\%\footnote{Due to the small number of live samples, thresholds at lower False Detection Rate (FDR) such as 0.2\% (recommended under the IARPA ODIN program) cannot be computed.} False Detection Rate (FDR) for our evaluation. 

Given a confusion matrix with the number of True Positives ($TP$), True Negatives ($TN$), False Positives ($FP$), and False Negatives ($FN$), we first define the Attack Presentation Classification Error Rate (APCER) and the Bonafide Presentation Classification Error Rate (BPCER)\footnote{APCER corresponds to the worst error rate among the spoof types (akin to False Rejection Rate) and the BPCER is the error in classifying lives as spoofs (akin to False Detection Rate).} as,
\begin{align}
    APCER &= \frac{FN}{FN+TP},\\
    BPCER &= \frac{FP}{FP+TN}
\end{align}

The ACER is then computed by,
\begin{align}
    ACER &= \frac{\underset{\rm k = 1\ldots C}{\max}(APCER_{k}) + BPCER}{2}
\end{align}
where $C$ is the number of spoof types (provided in Table~\ref{tab:public_datasets}).

The Equal Error Rate is the error defined at a threshold where both False Detection Rate and False Rejection Rate is minimal. For cross-dataset evaluation, the Half Total Error Rate (HTER) is used,
\begin{align}
    HTER &= \frac{FDR+FRR}{2}
\end{align}

\begin{table}[!t] 
\footnotesize
\caption{Generalization error (EER \%) on learning global (CNN) vs. local (FCN) representations of SiW-M~\cite{deep_tree}.}
\centering
\def\arraystretch{1.5}
\begin{tabular}{c||c|c|c|c}
\noalign{\hrule height 1.5pt}
\textbf{Method} & \textbf{Replay} & \textbf{Obfuscation} & \textbf{Paper Glasses} & \textbf{Overall}\\ \hline
CNN & 12.8 & 44.6 & 23.6 & 27.0 $\pm$ 13.2\\ \hline
FCN & 11.2 & 37.6 & 12.4 & 20.4 $\pm$ 12.1\\
\hline
\noalign{\hrule height 1.5pt}
\end{tabular}
\label{tab:cnn_v_fcn}
\end{table}


\begin{figure}[!t]
    \centering
    \subfloat[Manually-defined Regions]{\includegraphics[width=0.8\linewidth]{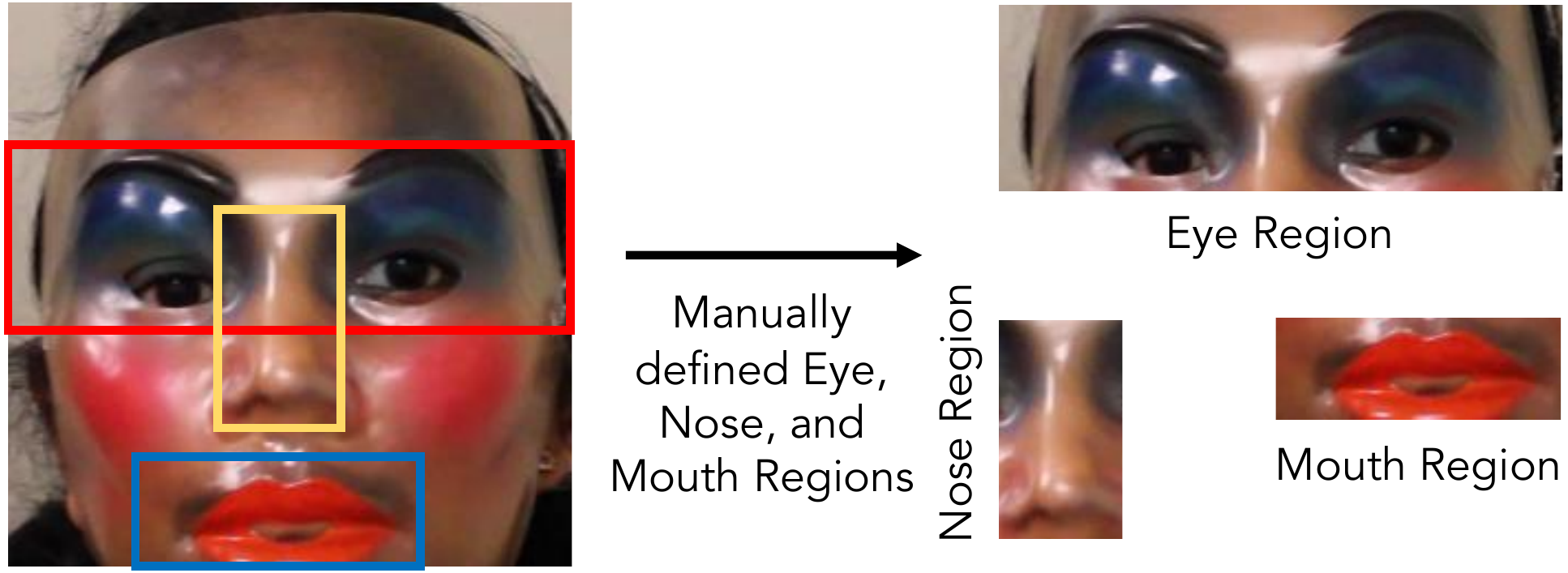}}\\
    \subfloat[Landmark Based]{\includegraphics[width=0.8\linewidth]{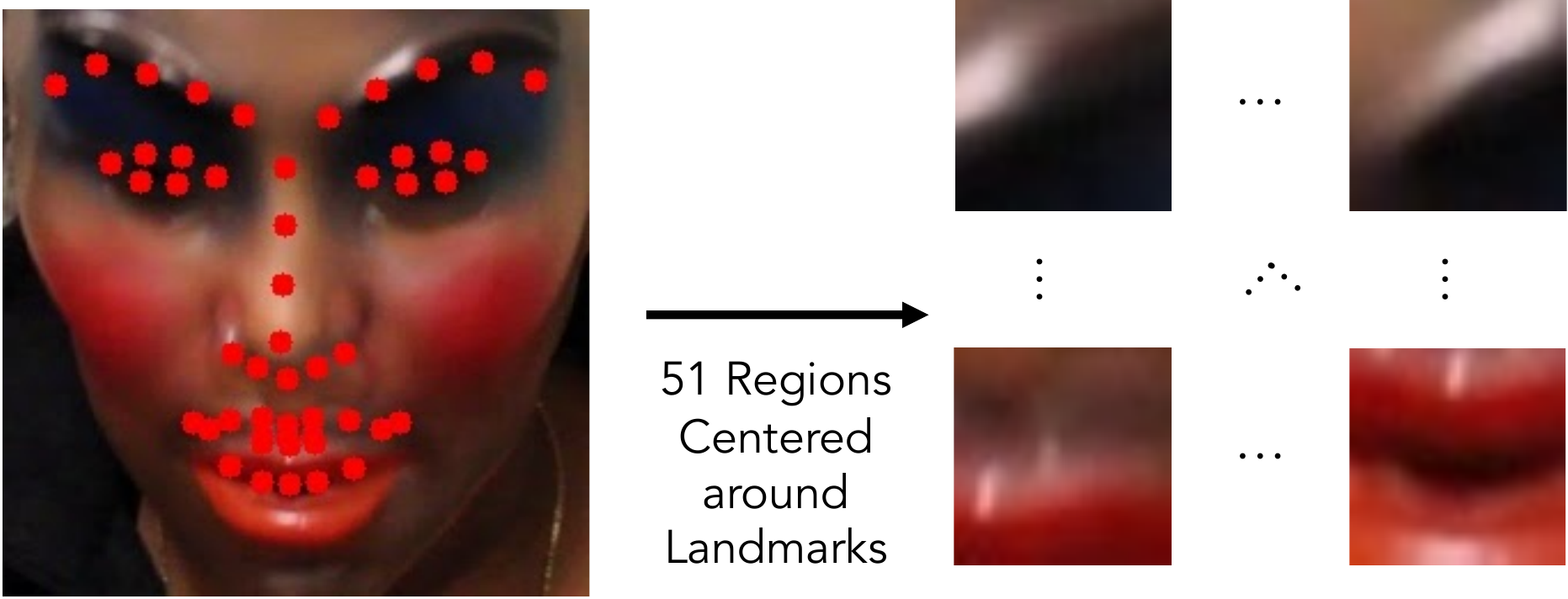}}\\
    \subfloat[Self-Supervised (Proposed)]{\includegraphics[width=0.8\linewidth]{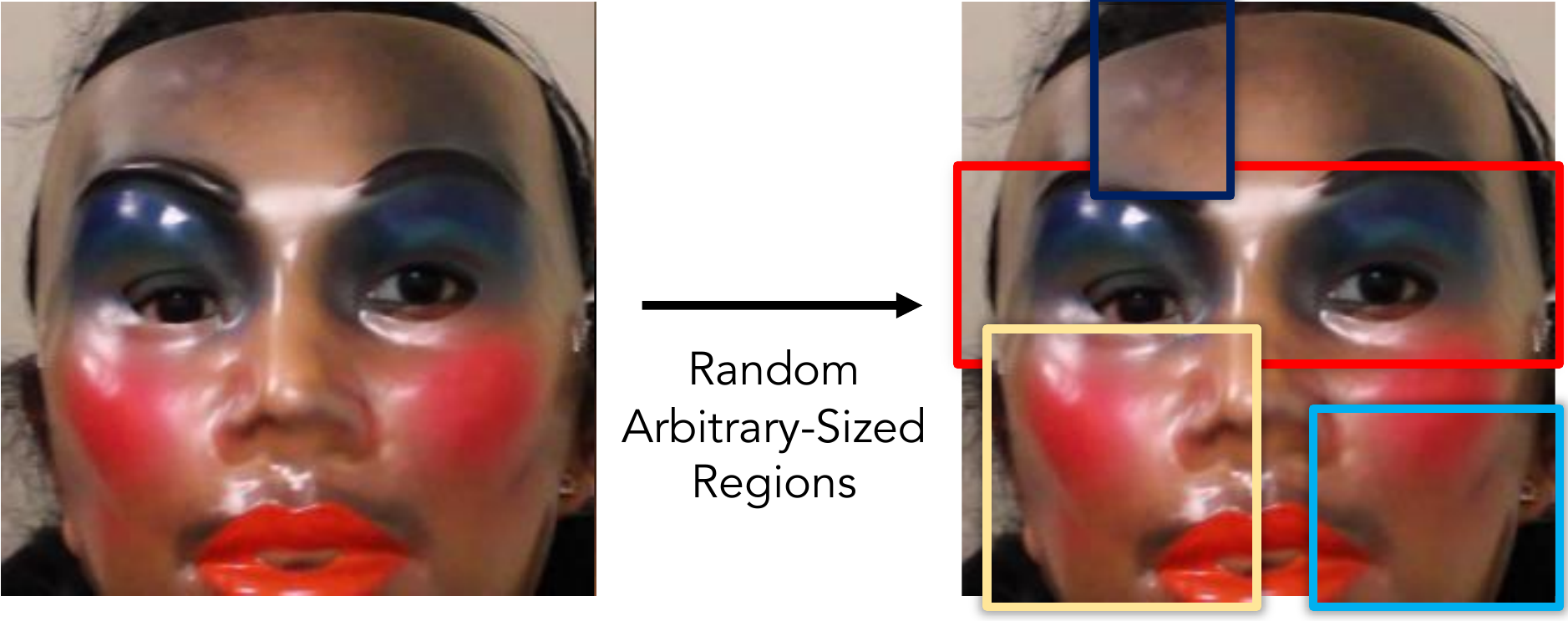}}
    \caption{Illustration of various region extraction strategies from training images. (a) and (b) are regions extracted via domain knowledge (manually defined) or landmark-based. (c) random regions extracted via proposed self-supervision scheme. Each color denotes a separate region.}
    \label{fig:region_extraction}
\end{figure}

\begin{table*}[!h] 
\footnotesize
\caption{Generalization performance of different region extraction strategies on SiW-M dataset. Here, each column represents an unknown spoof type while the method is trained on the remaining 12 spoof types.}
\centering
\def\arraystretch{1.5}
\resizebox{\textwidth}{!}{\begin{tabular}{c||c||c||c||c|c|c|c|c||c|c|c||c|c|c||c}
\noalign{\hrule height 1.5pt}
\multirow{7}{*}{\textbf{Method}} & \multirow{7}{*}{\textbf{Metric (\%)}} & \textbf{Replay} & \textbf{Print} & \multicolumn{5}{c||}{\textbf{Mask Attacks}} &\multicolumn{3}{c||}{\textbf{Makeup Attacks}} &\multicolumn{3}{c||}{\textbf{Partial Attacks}} & \multirow{7}{*}{\textbf{Mean $\pm$ Std.}}\\ \hline
& & {\includegraphics[height=1cm]{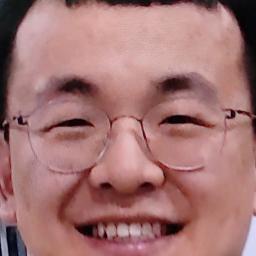}} & {\includegraphics[height=1cm]{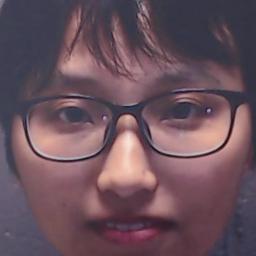}} & {\includegraphics[height=1cm]{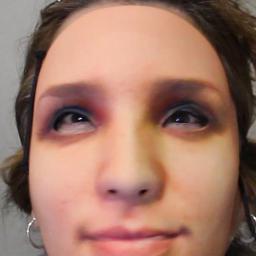}} & {\includegraphics[height=1cm]{Images/collage/silicone.jpg}} & {\includegraphics[height=1cm]{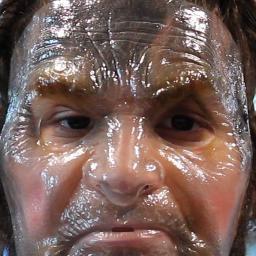}} & {\includegraphics[height=1cm]{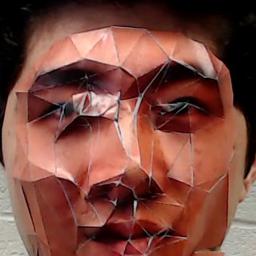}} & {\includegraphics[height=1cm]{Images/collage/mann4.jpg}} & {\includegraphics[height=1cm]{Images/collage/obsufcation.jpg}} & {\includegraphics[height=1cm]{Images/collage/makeup3.jpg}} & {\includegraphics[height=1cm]{Images/collage/cosmetic.jpg}} & {\includegraphics[height=1cm]{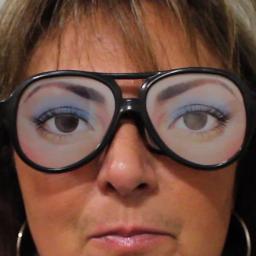}} & {\includegraphics[height=1cm]{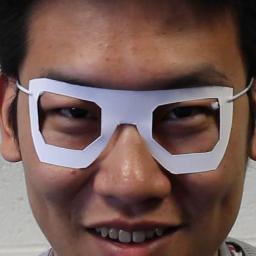}} & {\includegraphics[height=1cm]{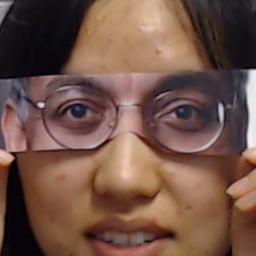}} &\\
& & \textbf{Replay} & \textbf{Print} & \textbf{Half} & \textbf{Silicone} & \textbf{Trans.} & \textbf{Paper} & \textbf{Mann.} & \textbf{Obf.} & \textbf{Imp.} & \textbf{Cosm.} & \textbf{FunnyEye} & \textbf{Glasses} & \textbf{Paper Cut} &\\
& & 99 vids. & 118 vids. & 72 vids. & 27 vids. & 88 vids. & 17 vids. & 40 vids. & 23 vids.& 61 vids. & 50 vids. & 160 vids. & 127 vids. & 86 vids.\\
\hline\hline
\multirow{2}{*}{Global} 
& ACER & 11.2 & 15.5 & 12.8 & 21.5 & 35.4 & 6.1 & 10.7 & 52.2 & 50.0 & 20.5 & 26.2 & 12.1 & 9.6 & 22.6 $\pm$ 15.3\\
& EER &  11.2 & 14.0 & 12.8 & 23.1 & 26.6 & 2.9 & 11.0 & 37.6 & 10.4 & 17.0 & 24.2 & 12.4 & 10.1 & 16.8 $\pm$ 9.3\\
\noalign{\hrule height 0.8pt}
\multirow{2}{*}{Eye-Region} 
& ACER &  13.2 & 13.7 & 7.5 & 17.4 & 22.5 & 5.79 & 6.2 & 19.5 & 8.3 & 11.7 & 32.8 & 15.3 & 7.3 & 13.2 $\pm$ 8.5\\
& EER &  12.4 & 11.4 & 7.3 & 15.2 & 21.5 & 2.9 & 6.5 & 20.2 & 7.8 & 11.2 & 27.2 & 14.7 & 7.5 & 12.3 $\pm$ 6.2\\
\noalign{\hrule height 0.8pt}
\multirow{2}{*}{Nose-Region} 
& ACER &  17.4 & 10.5 & 8.2 & 13.8 & 30.3 & 5.3 & 8.4 & 37.4 & 5.1 & 18.0 & 35.5 & 31.4 & 7.1 & 17.6 $\pm$ 12.0\\
& EER &  14.6 & 9.8 & 9.2 & 12.7 & 22.0 & 5.2 & 8.4 & 23.6 & 4.4 & 14.6 & 24.9 & 27.7 & 7.6 & 14.2 $\pm$ 7.9\\
\noalign{\hrule height 0.8pt}
\multirow{2}{*}{Mouth-Region} 
& ACER &  20.5 & 20.7 & 22.9 & 26.3 & 30.6 & 15.6 & 17.1 & 44.2 & 18.1 & 24.0 & 38.0 & 47.2 & 8.5 & 25.7 $\pm$ 11.4\\
& EER &  19.9 & 21.3 & 22.6 & 25.1 & 30.0 & 10.1 & 10.7 & 40.9 & 16.1 & 24.0 & 35.5 & 40.4 & 8.1 & 23.4 $\pm$ 10.9\\
\noalign{\hrule height 0.8pt}
\multirow{2}{*}{Global + Eye + Nose} 
& ACER &  10.9 & 10.5 & 7.5 & 17.7 & 28.7 & 5.1 & 7.0 & 38.0 & 5.1 & 13.6 & 29.4 & 15.2 & 6.2 & 15 $\pm$ 10.7\\
& EER &  10.2 & 10.0 & 7.7 & 15.8 & 21.3 & 1.8 & 6.7 & 21.0 & 3.0 & 12.3 & 22.5 & 12.3 & 6.5 & 11.6 $\pm$ 6.8\\
\noalign{\hrule height 0.8pt}
\multirow{2}{*}{Landmark-Region} 
& ACER &  10.7 & 9.2 & 18.4 & 25.1 & 26.4 & 6.2 & 6.9 & 53.8 & 8.1 & 15.4 & 35.8 & 40.8 & 7.6 & 20.3 $\pm$ 15.2\\
& EER &  8.0 & 10.1 & 12.2 & 23.1 & 18.8 & 8.9 & 4.1 & 40.1 & 9.9 & 15.6 & 17.7 & 25.6 & 4.9 & 15.3 $\pm$ 10\\
\noalign{\hrule height 0.8pt}
\multirow{2}{*}{Global + Landmark} 
& ACER &   12.0 & 11.2 & 7.3 & 23.7 & 26.4 & 6.3 & 5.9 & 26.7 & 6.7 & 10.7 & 27.8 & 25.7 & 6.4 & 15.1 $\pm$ 9.2\\
& EER & 11.5 & 10.1 & 7.2 & 19.0 & 4.9 & 6.6 & 4.6 & 25.6 & 6.7 & 10.9 & 23.5 & 18.5 & 4.7 & 11.8 $\pm$ 7.4\\
\noalign{\hrule height 0.8pt}
\multirow{2}{*}{Self-Sup. (Proposed)}
& ACER &  7.4 & 19.5 & 3.2 & 7.7 & 33.3 & 5.2 & 3.3 & 22.5 & 5.9 & 11.7 & 21.7 & 14.1 & 6.4 & \textbf{12.4 $\pm$ 9.2}\\
& EER &  6.8 & 11.2 & 2.8 & 6.3 & 28.5 & 0.4 & 3.3 & 17.8 & 3.9 & 11.7 & 21.6 & 13.5 & 3.6 & \textbf{10.1 $\pm$ 8.4}\\
\hline
\noalign{\hrule height 1.5pt}
\end{tabular}}
\label{tab:ablation_regions}
\end{table*}

\subsection{Evaluation of Global Descriptor vs. Local Representation}
 In order to analyze the impact of learning local embeddings as opposed to learning a global embedding, we conduct an ablation study on three spoof types in the SiW-M dataset~\cite{deep_tree}, namely, Replay (Figure~\ref{fig:spoof_types}c), Obfuscation (Figure~\ref{fig:spoof_types}i), and Paper Glasses (Figure~\ref{fig:spoof_types}m). 

In this experiment, a~\emph{traditional CNN} learning a global image descriptor is constructed by replacing the $1\times 1$ convolutional layer with a fully connected layer. We compare the CNN to the proposed backbone \emph{FCN} in Table~\ref{tab:params} which learns local representations. For a fair comparison between CNN and FCN, we utilize the same meta-parameters and employ global supervision only (Stage I).

In Table~\ref{tab:cnn_v_fcn}, we find that overall FCNs are more generalizable to unknown spoof types compared to global embeddings. For spoof types where spoof affects the entire face, such as replay attacks, the differences between generalization performance of CNN and FCN are negligible. Here, spoof decisions at local spatial regions does not have any significant advantage over a single spoof decision over the entire image. Recall that CNNs employ a fully connected layer which strips away all spatial information. This explains why local decisions can significantly improve generalizability of FCN over CNN when spoof types are local in nature (\eg, make-up attacks and partial attacks). Due to subtlety of obfuscation attacks and localized nature of paper glasses, FCN can exhibit a relative reduction in EER by $16\%$ and $47\%$, respectively, relative to CNN.

\subsection{Region Extraction Strategies}
  We considered 5 different region extraction strategies, namely,~\emph{Eye-Region},~\emph{Nose-Region},~\emph{Mouth-Region},~\emph{Landmark-Region}, and~\emph{Self-Supervised Regions (Proposed)} (see Figure~\ref{fig:region_extraction}). Here, \emph{Global} refers to training the FCN with the entire face image only (Stage I).
 
 Since all face images are aligned and cropped, spatial information is consistent across all images in datasets. Therefore, we can automatically extract facial regions that inclue eye, nose, and mouth regions (Figure~\ref{fig:region_extraction}a). We train the proposed FCN separately on each of the three regions to obtain three models: eye-region, nose-region, and mouth-region.
 
 We also investigate extracting regions defined by face landmarks. For this, we utilize a state-of-the-art landmark extractor, namely DLIB~\cite{dlib}, to obtain 51 landmark points around eyebrows, eyes, nose, and mouth. A total of 51 regions (with a fixed size $32\times 32$) centered around each landmark are extracted and used train a single FCN on all 51 regions.
 
 Our findings are as follows: (i) almost all methods with regional supervision have lower overall error rates as compared to training with the entire face. Exception to this is when FCN is trained only on mouth regions. This is likely because a majority of spoof types may not contain spoof patterns across mouth regions (Figure~\ref{fig:spoof_types}). (ii) when both global and domain-knowledge strategies (specifically, eyes and nose) are fused, the generalization performance improves compared to the global model alone. Note that we do not fuse the mouth region since the performance is poor for mouth regions. Similarly, we find that regions cropped around landmarks when fused with the global classifier can achieve better generalization performance. (iii) compared to all region extraction strategies, the proposed self-supervised region extraction strategy (Stage II) achieves the lowest generalization error rates across all spoof types with a $40\%$ and $45\%$ relative reduction in EER and ACER compared to the Global model (Stage I). This supports our hypothesis that both Stage I and Stage II are required for enhanced generalization performance across unknown spoof types. A score-level fusion of the global FCN with self-supervised regions does not show any significant reduction in error rates. This is because we already trained the proposed FCN on global faces in Stage I.
 
  \begin{figure}[!t]
    \centering
    \subfloat[Obfuscation]{\includegraphics[width=0.32\linewidth]{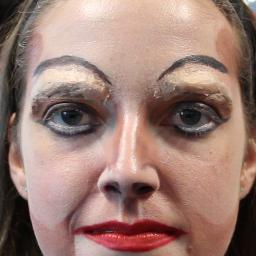}}\hfill
    \subfloat[Spoofness Score: 1.0]{\includegraphics[width=0.32\linewidth]{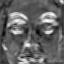}}\hfill
    \subfloat[Spoofness Score: 0.7]{\includegraphics[width=0.32\linewidth]{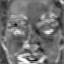}}\\
    \subfloat[Obfuscation]{\includegraphics[width=0.32\linewidth]{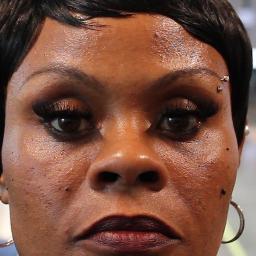}}\hfill
    \subfloat[Spoofness Score: 0.0]{\includegraphics[width=0.32\linewidth]{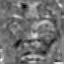}}\hfill
    \subfloat[Spoofness Score: 0.0]{\includegraphics[width=0.32\linewidth]{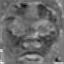}}\\
    \caption{(a) An example obfuscation spoof attempt where our network correctly predicts the input to be a spoof. (b, e) Score map output by our network trained via Self-Supervised Regions. (c, f) Score map output by FCN trained on entire face images. (d) An example obfuscation spoof attempt where our network \emph{incorrectly} predicts the input to be a live. Spoofness scores are given below the spoof maps. Decision threshold is $0.5$.}
    \label{fig:spoof_analysis}
\end{figure}
 
 \begin{table}[!t] 
\footnotesize
\caption{Generalization error (EER $\%$) of FCNs with respect to the number of trainable parameters.}
\centering
\def\arraystretch{1.5}
\resizebox{\linewidth}{!}{\begin{tabular}{c||c|c|c|c}
\noalign{\hrule height 1.5pt}
\textbf{Method} & \textbf{Replay} & \textbf{Obfuscation} & \textbf{Paper Glasses} & \textbf{Mean $\pm$ Std.}\\ \hline
3-layers ($76K$) & 14.0 & 44.1 & 7.5 & 28.5 $\pm$ 12.3\\ \hline
5-layers ($1.5M$; proposed) & \textbf{6.8} & \textbf{17.8} & \textbf{23.6} & \textbf{12.7 $\pm$ 4.5}\\ \hline
6-layers ($3M$) & 7.8 & 25.19 & 19.7 & 17.6 $\pm$ 7.2\\ \hline
\noalign{\hrule height 1.5pt}
\end{tabular}}
\label{tab:capacity}
\end{table}

\begin{table*}[!h] 
\footnotesize
\caption{Results on SiW-M: Unknown Attacks. Here, each column represents an unknown spoof type while the method is trained on the remaining 12 spoof types.}
\centering
\def\arraystretch{1.5}
\resizebox{\textwidth}{!}{\begin{tabular}{c||c||c||c||c|c|c|c|c||c|c|c||c|c|c||c}
\noalign{\hrule height 1.5pt}
\multirow{7}{*}{\textbf{Method}} & \multirow{7}{*}{\textbf{Metric}} & \textbf{Replay} & \textbf{Print} & \multicolumn{5}{c||}{\textbf{Mask Attacks}} &\multicolumn{3}{c||}{\textbf{Makeup Attacks}} &\multicolumn{3}{c||}{\textbf{Partial Attacks}} & \multirow{7}{*}{\textbf{Mean $\pm$ Std.}}\\ \hline
& & {\includegraphics[height=1cm]{Images/collage/replay1.jpg}} & {\includegraphics[height=1cm]{Images/collage/paper2.jpg}} & {\includegraphics[height=1cm]{Images/collage/half_mask.jpg}} & {\includegraphics[height=1cm]{Images/collage/silicone.jpg}} & {\includegraphics[height=1cm]{Images/collage/mask1.jpg}} & {\includegraphics[height=1cm]{Images/collage/paper_mask.jpg}} & {\includegraphics[height=1cm]{Images/collage/mann4.jpg}} & {\includegraphics[height=1cm]{Images/collage/obsufcation.jpg}} & {\includegraphics[height=1cm]{Images/collage/makeup3.jpg}} & {\includegraphics[height=1cm]{Images/collage/cosmetic.jpg}} & {\includegraphics[height=1cm]{Images/collage/partial_funnyeye.jpg}} & {\includegraphics[height=1cm]{Images/collage/partial_glass.jpg}} & {\includegraphics[height=1cm]{Images/collage/partial_eye.jpg}} &\\
& & \textbf{Replay} & \textbf{Print} & \textbf{Half} & \textbf{Silicone} & \textbf{Trans.} & \textbf{Paper} & \textbf{Mann.} & \textbf{Obf.} & \textbf{Imp.} & \textbf{Cosm.} & \textbf{FunnyEye} & \textbf{Glasses} & \textbf{Paper Cut} &\\
& & 99 vids. & 118 vids. & 72 vids. & 27 vids. & 88 vids. & 17 vids. & 40 vids. & 23 vids.& 61 vids. & 50 vids. & 160 vids. & 127 vids. & 86 vids.\\
\hline\hline
\multirow{2}{*}{SVM+LBP~\cite{oulu_npu}} 
& ACER & 20.6 & 18.4 & 31.3 & 21.4 & 45.5 & 11.6 & 13.8 & 59.3 & 23.9 & 16.7 & 35.9 & 39.2 & 11.7 & 26.9 $\pm$ 14.5\\
& EER & 20.8 & 18.6 & 36.3 & 21.4 & 37.2 & 7.5 & 14.1 & 51.2 & 19.8 & 16.1 & 34.4 & 33.0 & 7.9 & 24.5 $\pm$ 12.9\\ \noalign{\hrule height 0.8pt}
\multirow{2}{*}{Auxiliary~\cite{siw}} 
& ACER & 16.8 & 6.9 & 19.3 & 14.9 & 52.1 & 8.0 & 12.8 & 55.8 & 13.7 & 11.7 & 49.0 & 40.5 & 5.3 & 23.6 $\pm$ 18.5\\
& EER & 14.0 & 4.3 & 11.6 & 12.4 & 24.6 & 7.8 & 10.0 & 72.3 & 10.1 & \textbf{9.4} & 21.4 & 18.6 & 4.0 & 17.0 $\pm$ 17.7\\ \noalign{\hrule height 0.8pt}
\multirow{2}{*}{DTN~\cite{deep_tree}} 
& ACER & 9.8 & \textbf{6.0} & 15.0 & 18.7 & 36.0 & 4.5 & 7.7 & 48.1 & 11.4 & 14.2 & \textbf{19.3} & 19.8 & 8.5 & 16.8 $\pm$ 11.1\\
& EER & 10.0 & \textbf{2.1} & 14.4 & 18.6 & 26.5 & 5.7 & 9.6 & 50.2 & 10.1 & 13.2 & \textbf{19.8} & 20.5 & 8.8 & 16.1 $\pm$ 12.2\\
\noalign{\hrule height 0.8pt}
\multirow{2}{*}{CDC~\cite{cdc}} 
& ACER & 10.8 & 7.3 & 9.1 & 10.3 & \textbf{18.8} & \textbf{3.5} & 5.6 & 42.1 & \textbf{0.8} &14.0 & 24.0 & 17.6 & \textbf{1.9} & 12.7 $\pm$ 11.2\\
& EER & 9.2 & 5.6 & 4.2 & 11.1 & \textbf{19.3} & 5.9 & 5.0 & 43.5 & \textbf{0.0} & 14.0 & 23.3 & 14.3 & \textbf{0.0} & 11.9 $\pm$ 11.8\\
\noalign{\hrule height 0.8pt}
\multirow{3}{*}{Proposed}
& ACER &  \textbf{7.4} & 19.5 & \textbf{3.2} & \textbf{7.7} & 33.3 & 5.2 & \textbf{3.3} & \textbf{22.5} & 5.9 & \textbf{11.7} & 21.7 & \textbf{14.1} & 6.4 & \textbf{12.4 $\pm$ 9.2}\\
& EER &  \textbf{6.8} & 11.2 & \textbf{2.8} & \textbf{6.3} & 28.5 & \textbf{0.4} & \textbf{3.3} & \textbf{17.8} & 3.9 & 11.7 & 21.6 & \textbf{13.5} & 3.6 & \textbf{10.1 $\pm$ 8.4}\\
\cmidrule[1.2pt]{2-16}
& TDR* &  72.0 & 51.0& 96.0 & 55.9 & 39.0 & 100.0 & 95.0 & 31.0 & 90.0 & 44.0 & 33.0 & 42.9 & 94.7 & \textbf{65.0 $\pm$ 25.9}\\
\noalign{\hrule height 1.5pt}
\end{tabular}}
\begin{tablenotes}
\item {\footnotesize \vspace{0.3em} *TDR evaluated at $2.0\%$ FDR}
\end{tablenotes}
\label{tab:results_siwm}
\end{table*}

\begin{table*}[!h] 
\footnotesize
\caption{Results on SiW-M: Known Spoof Types.}
\centering
\def\arraystretch{1.5}
\resizebox{\textwidth}{!}{\begin{tabular}{c||c||c||c||c|c|c|c|c||c|c|c||c|c|c||c}
\noalign{\hrule height 1.5pt}
\textbf{Method} & \textbf{Metric (\%)} &  &  & \multicolumn{5}{c||}{\textbf{Mask Attacks}} &\multicolumn{3}{c||}{\textbf{Makeup Attacks}} &\multicolumn{3}{c||}{\textbf{Partial Attacks}} & \textbf{Mean $\pm$ Std.}\\ \hline
& & \textbf{Replay} & \textbf{Print} & \textbf{Half} & \textbf{Silicone} & \textbf{Trans.} & \textbf{Paper} & \textbf{Mann.} & \textbf{Obf.} & \textbf{Imp.} & \textbf{Cosm.} & \textbf{Funny Eye} & \textbf{Glasses} & \textbf{Paper Cut} &\\
\hline\hline
\multirow{2}{*}{Auxiliary~\cite{siw}} 

& ACER &  5.1 & 5.0 & 5.0 & 10.2 & 5.0 & 9.8 & 6.3 & 19.6 & 5.0 & 26.5 & 5.5 & 5.2 & 5.0 & 8.7 $\pm$ 6.8\\
& EER &   4.7 & \textbf{0.0} & 1.6 & 10.5 & 4.6 & 10.0 & 6.4 & 12.7 & \textbf{0.0} & 19.6 & 7.2 & 7.5 & 0.0 & 6.5 $\pm$ 5.8\\
\noalign{\hrule height 0.8pt}
\multirow{3}{*}{Proposed}
& ACER &  \textbf{3.5} & \textbf{3.1} & \textbf{1.9} & \textbf{5.7} & \textbf{2.1} & \textbf{1.9} & \textbf{4.2} & \textbf{7.2} & \textbf{2.5} & \textbf{22.5} & \textbf{1.9} & \textbf{2.2} & \textbf{1.9} & \textbf{4.7 $\pm$ 5.6}\\
& EER &  \textbf{3.5} & 3.1 & \textbf{0.1} & \textbf{9.9} & \textbf{1.4} & \textbf{0.0} & \textbf{4.3} & \textbf{6.4} & 2.0 & \textbf{15.4} & \textbf{0.5} & \textbf{1.6} & \textbf{1.7} & \textbf{3.9 $\pm$ 4.4}\\
\cmidrule[1.2pt]{2-16}
& TDR*  &  55.5 & 92.3 & 69.5 & 100.0 & 90.4 & 100.0 & 85.1 & 92.5 & 78.7 & 99.1 & 95.6 & 95.7 & 76.0 & 87.0 $\pm$ 13.0\\ 
\noalign{\hrule height 1.5pt}
\end{tabular}}
\begin{tablenotes}
\item {\footnotesize \vspace{0.3em} *TDR evaluated at $2.0\%$ FDR}
\end{tablenotes}
\label{tab:results_siwm_known}
\end{table*}

 In Figure~\ref{fig:spoof_analysis}, we analyze the effect of training the FCN locally vs. globally on the prediction results. In the first row, where both models correctly predict the input to be a spoof, we see that FCN trained via random regions can correctly identify spoof regions such as fake eyebrows.  In contrast, the global FCN can barely locate the fake eyebrows. Since random regions increases the variability in the training set along with advantage of learning local features than global FCN, we find that proposed self-supervised regional supervision performs best.

\subsection{Evaluation of Network Capacity}
In Table~\ref{tab:capacity}, we plot the generalization performance of our model when we vary the capacity of the network. We consider three different variants of the proposed FCN: (a) 3-layer FCN ($76K$ parameters), (b) 5-layer FCN ($1.5$ parameters; proposed), and (c) 6-layer FCN ($3M$ parameters). This experiment is evaluated on three unknown spoof types, namely,~\emph{Replay},~\emph{Obfuscation}, and~\emph{Paper Glasses}. We chose these spoof types due to their vastly diverse nature. Replay attacks consist of global spoof patters, whereas obfuscation attacks are extremely subtle cosmetic changes. Paper Glasses are constrained only to eyes. While to many trainable parameters leads to poor generalization due to overfitting to the spoof types seen during training, whereas, too few parameters limits learning discriminative features. Based on this observation and experimental results, we utilize the 5-layer FCN (see Table~\ref{tab:params}) with approximately 1.5M parameters. A majority of prior studies employ 13 densely connected convolutional layers with trainable parameters exceeding $2.7M$~\cite{siw, cdc, pixel, sun2020face}.


\subsection{Generalization across Unknown Attacks}
The primary objective of this work is to enhance generalization performance across a multitude of unknown spoof types in order to effectively gauge the expected error rates in real-world scenarios. The evaluation protocol in SiW-M follows a leave-one-spoof-out testing protocol where the training split contains 12 different spoof types and the 13\textsuperscript{th} spoof type is held out for testing. Among the live videos, $80\%$ are kept in the training set and the remaining $20\%$ is used for testing lives. Note that there are no overlapping subjects between the training and testing sets. Also note that no data sample from the testing spoof type is used for validation since we evaluate our approach under unknown attacks. We report ACER and EER across the 13 splits. In addition to ACER and EER, we also report the TDR at 2.0\% FDR.

In Table~\ref{tab:results_siwm}, we compare \NAME with prior work. We find that our proposed method achieves significant improvement in comparison to the published results~\cite{cdc} (relative reduction of 14\% on the average EER and 3\% on the average ACER). Note that the standard deviation across all 13 spoof types is also reduced compared to prior approaches, even though some of them~\cite{cdc,siw} utilize auxiliary data such as depth and temporal information.

Specifically, we reduce the EERs of replay, half mask, transparent mask, silicone mask, paper mask, mannequin head, obfuscation, impersonation, and paper glasses relatively by 27\%, 33\%, 43\%, 93\%, 34\%, 59\%, and 6\%, respectively. Among all the 13 spoof types, detecting obfuscation attacks is the most challenging. This is due to the fact that the makeup applied in these attacks are very subtle and majority of the faces are live. Prior works were not successful in detectint these attacks and predict most of the obfuscation attacks as lives. By learning discriminative features locally, our proposed network improves the state-of-the-art obfuscation attack detection performance by 59\% in terms of EER and 46\% in terms of ACER. 

\subsection{SiW-M: Detecting Known Attacks}
Here all the 13 spoof types in SiW-M are used for training as well as testing. We randomly split the SiW-M dataset into a 60\%-40\% training/testing split and report the results in Table~\ref{tab:results_siwm_known}. In comparison to a state-of-the-art face anti-spoofing method~\cite{siw}, our method outperforms for almost all of the individual spoof types as well as the overall performance across spoof types. \emph{Auxiliary~\cite{siw}} utilizes depth and temporal information for spoof detection which adds significant complexity to the network.

\begin{table}[!t] 
\footnotesize
\caption{Error Rates (\%) of the proposed \NAME and and competing face spoof detectors under the four standard protocols of Oulu-NPU~\cite{oulu_npu}.}
\centering
\def\arraystretch{1.5}
\resizebox{\linewidth}{!}{\begin{tabular}{c||l|c|c|c}
\noalign{\hrule height 1.5pt}
 \textbf{Protocol} & \textbf{Method} & \textbf{APCER} & \textbf{BPCER} & \textbf{ACER}\\ \hline
 \multirow{5}{*}{I} & GRADIENT~\cite{competition} & 1.3 & 12.5 & 6.9\\
                    & Auxiliary~\cite{siw} & 1.6 & 1.6 & 1.6\\
                    & DeepPixBiS~\cite{pixel} & \textbf{0.8} & \textbf{0.0} & \textbf{0.4}\\
                    & TSCNN-ResNet~\cite{tscnn} & 5.1 & 6.7 & 5.9\\
                    & \NAME (Proposed) & 1.5  & 7.7  & 4.6\\ \hline
 \multirow{5}{*}{II} 
                    & GRADIENT~\cite{competition} & 3.1 & 1.9 & \textbf{2.5}\\
                    & Auxiliary~\cite{siw} & \textbf{2.7} & 2.7 & 2.7\\
                    & DeepPixBiS~\cite{pixel} &11.4 & \textbf{0.6} & 6.0\\
                    & TSCNN-ResNet~\cite{tscnn} & 7.6 & 2.2 & 4.9\\
                    & \NAME (Proposed) & 3.1 & 3.7 & 3.4 \\ \hline
 \multirow{5}{*}{III} 
                    & GRADIENT~\cite{competition} & \textbf{2.1 $\pm$ 3.9} & $5.0\pm 5.3$ & $3.8 \pm 2.4$\\
                    & Auxiliary~\cite{siw} & $2.7\pm 1.3$ & $3.1\pm 1.7$ & $2.9 \pm 1.5$\\
                    & DeepPixBiS~\cite{pixel} &$11.7\pm 19.6$ & $10.6\pm 14.1$ & $11.1 \pm 9.4$\\
                    & TSCNN-ResNet~\cite{tscnn} & $3.9\pm 2.8$ & $7.3 \pm 1.1$ & $5.6 \pm 1.6$\\
                    & \NAME (Proposed) &  2.9 $\pm$ 2.1 & \textbf{2.7 $\pm$ 3.2}  & \textbf{2.8 $\pm$ 2.2} \\ \hline
 \multirow{5}{*}{IV} & GRADIENT~\cite{competition} & \textbf{5.0 $\pm$ 4.5} & $15.0\pm 7.1$ & $10.0 \pm 5.0$\\
                    & Auxiliary~\cite{siw} & $9.3\pm 5.6$ & $10.4\pm 6.0$ & \textbf{9.5 $\pm$ 6.0}\\
                    & DeepPixBiS~\cite{pixel} &$36.7\pm 29.7$ & $13.3\pm 16.8$ & $25.0 \pm 12.7$\\
                    & TSCNN-ResNet~\cite{tscnn} & $11.3\pm 3.9$ & \textbf{9.7 $\pm$ 4.8} & $9.8 \pm 4.2$\\
                    & \NAME (Proposed) & $8.3 \pm 6.8$  & $13.3 \pm 8.7$ & $10.8 \pm 5.1$ \\
\hline\hline
\noalign{\hrule height 1.5pt}
\end{tabular}}
\label{tab:oulu}
\end{table}

\subsection{Evaluation on Oulu-NPU Dataset}
We follow the four standard protocols defined in the OULU-NPU dataset~\cite{oulu_npu} which cover the cross-background, cross-presentation-attack-instrument  (cross-PAI),  cross-capture-device,  and cross-conditions evaluations: 
\begin{itemize}
    \item \textbf{Protocol I:} unseen subjects, illumination, and backgrounds;
    \item \textbf{Protocol II:} unseen subjects and attack devices;
    \item \textbf{Protocol III:} unseen subjects and cameras;
    \item \textbf{Protocol IV:} unseen subjects, illumination, backgrounds, attack devices, and cameras.
\end{itemize}

We compare the proposed \NAME with the best performing method, namely GRADIENT~\cite{competition}, in IJCB Mobile Face Anti-Spoofing Competition~\cite{competition} for each protocol.
We also include some newer baseline methods, including Auxiliary~\cite{siw}, DeepPixBiS~\cite{pixel}, and TSCNN~\cite{tscnn}. We compare our proposed method with 10 baselines in total for each protocol. Additional baselines can be found in supplementary material.

In Table~\ref{tab:oulu}, \NAME achieves ACERs of 4.6\%, 3.4\%, 2.8\%, and 10.8\% in the four protocols, respectively. Among the baselines, \NAME even outperforms prevailing state-of-the-art methods in protocol III which corresponds to generalization performance for unseen subjects and cameras. The results are comparable to baseline methods in the other three protocols. Since Oulu-NPU comprises of only print and replay attacks, a majority of the baseline methods incorporate auxiliary information such as depth and motion. Indeed, incorporating auxiliary information could improve the results at the risk of overfitting and overhead cost and time.

\begin{table}[!t] 
\footnotesize
\caption{Cross-Dataset HTER (\%) of the proposed \NAME and competing face spoof detectors.}
\centering
\def\arraystretch{1.2}
\resizebox{\linewidth}{!}{\begin{tabular}{l||c||c}
\noalign{\hrule height 1.5pt}
\textbf{Method} &  \textbf{CASIA} $\rightarrow$ \textbf{Replay} & \textbf{Replay} $\rightarrow$ \textbf{CASIA}\\
 \noalign{\hrule height 1.2pt}
CNN~\cite{yang} & 48.5  & 45.5 \\
Color Texture~\cite{lbp4} & 47.0  & 49.6 \\
Auxiliary~\cite{siw} & 27.6  &  \textbf{28.4}\\
De-Noising~\cite{despoof} & 28.5  & 41.1 \\
STASN~\cite{stasn} & 31.5 & 30.9\\
SAPLC~\cite{sun2020face} & 27.3 & 37.5 \\
\NAME (Proposed) & \textbf{19.9} & 41.9 \\
\noalign{\hrule height 1.5pt}
\end{tabular}}
\begin{tablenotes}
\item {\footnotesize \vspace{0.3em} ``CASIA $\rightarrow$ Replay" denotes training on CASIA and testing on Replay-Attack}
\end{tablenotes}
\label{tab:casia_replay}
\end{table}

\subsection{Cross-Dataset Generalization} In order to evaluate the generalization performance of \NAME when trained on one dataset and tested on another, following prior studies, we perform a cross-dataset experiment between CASIA-FASD~\cite{casia_mfsd} and Replay-Attack~\cite{replay_attack}. 

In Table~\ref{tab:casia_replay}, we find that, compared to 6 prevailing state-of-the-art methods, the proposed \NAME achieves the lowest error (a $27\%$ improvement in HTER) when trained on CASIA-FASD~\cite{casia_mfsd} and evaluated on Replay-Attack~\cite{replay_attack}.  On the other hand, \NAME achieves worse performance when trained on Replay-Attack and tested on CASIA-FASD. This can likely be attributed to higher resolution images in CASIA-FASD compared to Replay-Attack. This demonstrates that \NAME trained with higher-resolution data can generalize better on poorer quality testing images, but the reverse may not hold true. We intend on addressing this limitation in future work.

Additional baselines can be found in supplementary material.

\subsection{Failure Cases}
\begin{figure}[!t]
    \centering
    \includegraphics[width=\linewidth]{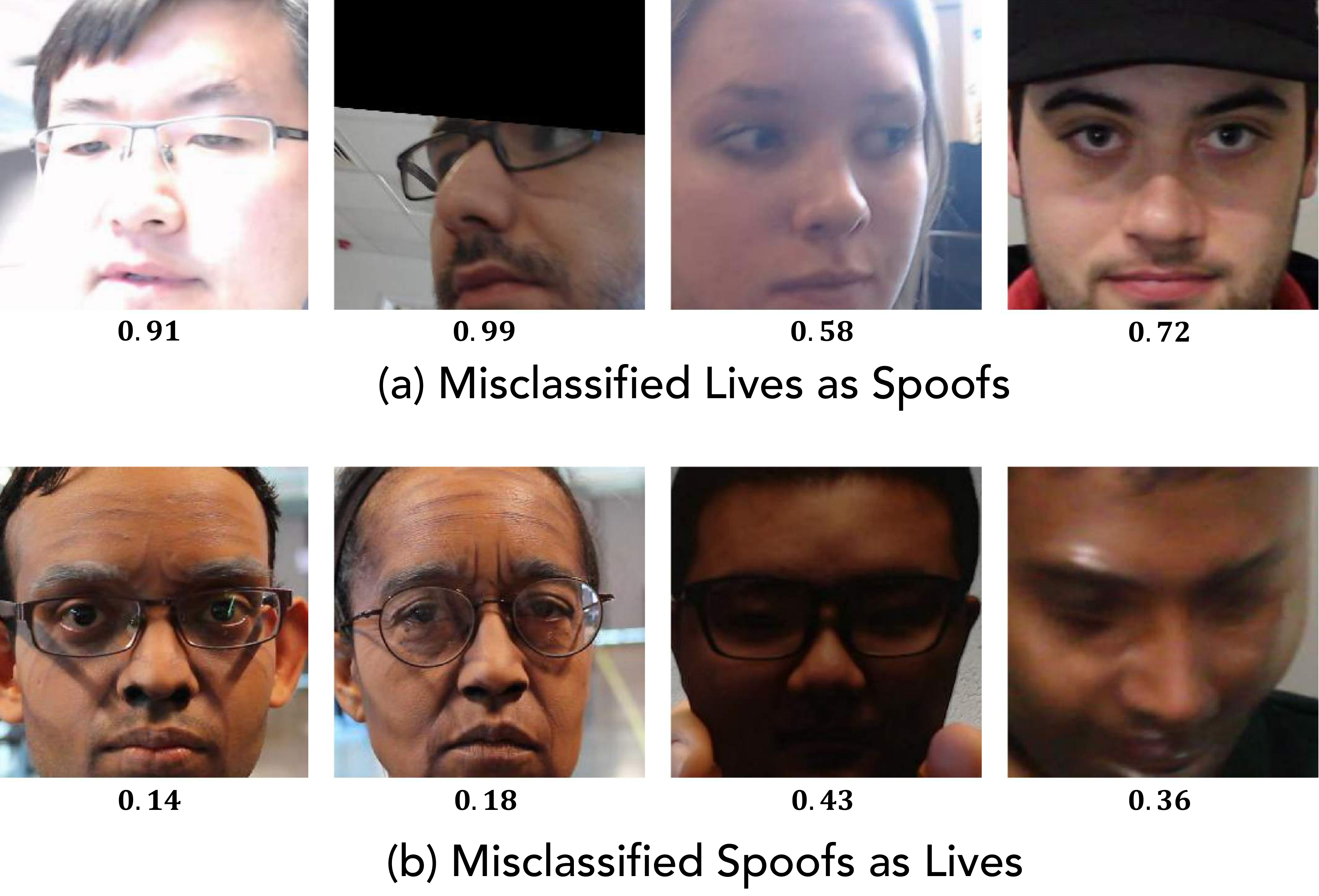}
    \caption{Example cases where the proposed framework, \NAMENOSPACE, fails to correctly classify lives and spoofs. (a) Lives are misclassified as spoofs likely due to bright lighting and occlusions in face regions. (b) Spoofs misclassified as lives due to the subtle nature of make-up attacks and transparent masks. Corresponding spoofness scores ($\in[0, 1]$) are provided below each image. Larger value of spoofness score indicates a higher likelihood that the input image is a spoof. Decision threshold is $0.5$.}
    \label{fig:misclassify}
\end{figure}

Even though experiment results show enhanced generalization performance, our model still fails to correctly classify certain input images. In Figure~\ref{fig:misclassify}, we show a few such examples. 

\begin{figure*}[!t]
    \centering
    \includegraphics[width=\linewidth]{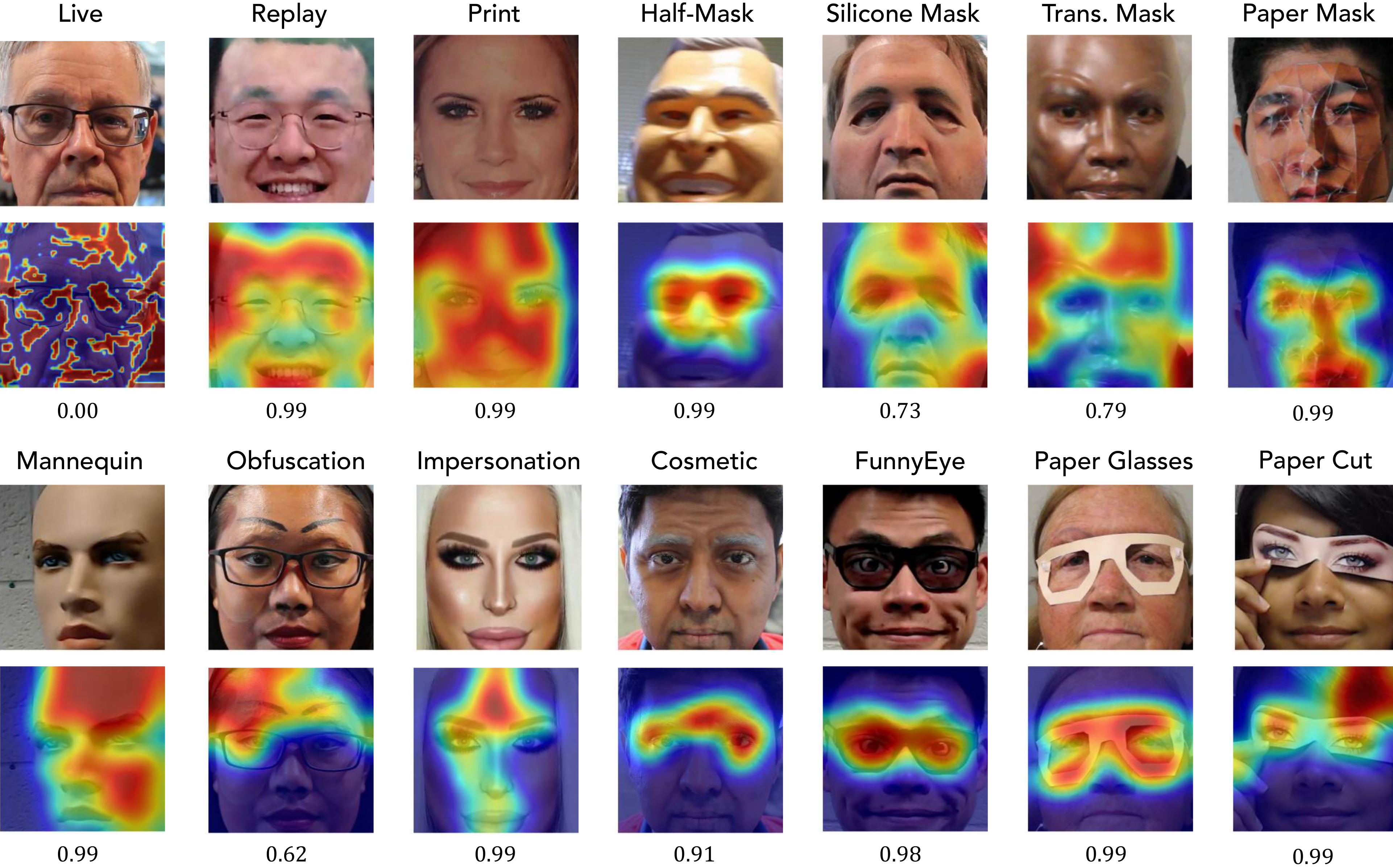}
    \caption{Visualizing spoof regions via the proposed~\NAMENOSPACE. Red regions indicate higher likelihood of being a spoof region. Corresponding spoofness scores ($\in[0, 1]$) are provided below each image. Larger value of spoofness score indicates a higher likelihood that the input image is a spoof. Decision threshold is $0.5$.}
    \label{fig:viz}
\end{figure*}

Figure~\ref{fig:misclassify}a shows incorrect prediction of lives as spoofs in the presence of inhomogeneous illumination. This is because the model predicts lives as being one of replay and print attacks which exhibit bright lighting patterns due to the recapturing media such as smartphones and laptops. Since we fine-tune our network via regional supervision in Stage II, artifacts which obstruct parts of the faces can also adversely affect our model.

Figure~\ref{fig:misclassify}b shows incorrect classification of spoofs as lives. This is particularly true when  suffers when spoof patterns are very subtle, such as cosmetic and obfuscation make-up attacks. Transparent masks can also be problematic when the mask itself is barely visible. 

\subsection{Computational Requirement}
Since face anti-spoofing modules are employed as a pre-processing step for automated face recognition systems, it is crucial that the spoof prediction time should be as low as possible. The proposed \NAME takes under $2$ hours to train both Stage I and Stage II, and $4$ miliseconds to predict a single $(256\times 256)$ spoof/live image on a Nvidia GTX 1080Ti GPU. In other words, \NAME can process frames at 250 Frames Per Second (FPS) and the size of the model is only $11.8$MB. Therefore, \NAME is well suited for deployment on embedded devices such as smartphones.

\subsection{Visualizing Spoof Regions}
\NAME can automatically locate the individual spoof regions in an input face image. In Figure~\ref{fig:viz}, we show heatmaps from the score maps extracted for a randomly chosen image from all spoof types. Red regions indicate a higher likelihood of spoof.

For a live input image, the spoof regions are sparse with low likelihoods. In the case of replay and print attacks, the predicted spoof regions are located throughout the entire face image. This is because these spoofs contain global-level noise. For mask attacks, including half-mask, silicone mask, transparent mask, paper mask, and mannequin, the spoof patterns are identified near the eye and nose regions. Make-up attacks are harder to detect since they are very subtle in nature. Proposed \NAME detects obfuscation and cosmetic attack attempts by learning local discriminative cues around the eyebrow regions. In contrast, impersonation make-up patterns exist throughout the entire face. We also find that \NAME can precisely locate the spoofing artifacts, such as funny eyeglasses, paper glasses, and paper cut, in partial attacks.

\section{Discussion}
We show that the proposed \NAME achieves superior generalization performance on SiW-M dataset~\cite{deep_tree} compared to the prevailing state-of-the-art methods that tend to overfit on the seen spoof types. Our method also achieves comparable performance to the state-of-the-art in Oulu-NPU dataset\cite{oulu_npu} and outperforms all baselines for cross-dataset generalization performance (CASIA-FASD~\cite{casia_mfsd} $\rightarrow$ Replay-Attack~\cite{replay_attack}). 

In contrast to a number of prior studies~\cite{siw, cdc, lbp4, patch}, the proposed approach does not utilize auxiliary cues for spoof detection, such as motion and depth information. While incorporating such cues may enhance performance on print and replay attack datasets such as Oulu-NPU, CASIA-MFSD, and Replay-Attack, it is at the risk of potentially overfitting to the two attacks and compute cost. A major benefit of \NAME lies is its usability. A simple pre-processing step includes face detection and alignment. The cropped face is then passed to the network. With a~\emph{single} forward-pass through the FCN, we obtain both the score map and the spoofness score.

Even though the proposed method is well-suited for generalizable face anti-spoofing, \NAME is still limited by the amount and quality of available training data. For instance, when trained on a low-resolution dataset, namely Replay-Attack~\cite{replay_attack}, cross-dataset generalization performance suffers.

\section{Conclusion}
Face anti-spoofing systems are crucial for secure operation of an automated face recognition system. With the introduction of sophisticated spoofs, such as high resolution and tight fitting silicone 3D face masks, spoof detectors need to be robust and generalizable. We proposed a face anti-spoofing framework, namely~\NAMENOSPACE, that achieved state-of-the-art generalization performance against 13 different spoof types. \NAME reduced the average error rate of competitive algorithms by $14\%$ on one of the largest and most diverse face anti-spoofing dataset, SiW-M, comprised of 13 spoof types. It also generalizes well when training and testing datasets are from different sources. In addition, the proposed method is shown to be more interpretable compared to prior studies since it can directly predict which parts of the faces are considered as spoofs. In the future, we intend on exploring whether incorporating domain knowledge in \NAME can further improve generalization performance.

\section*{Acknowledgment}
This research is based upon work supported in part by the Office of the Director of National Intelligence (ODNI), Intelligence Advanced Research Projects Activity (IARPA), via IARPA R\&D Contract No. $2017-17020200004$. The views and conclusions contained herein are those of the authors and should not be interpreted as necessarily representing the official policies, either expressed or implied, of ODNI, IARPA, or the U.S. Government. The U.S. Government is authorized to reproduce and distribute reprints for governmental purposes notwithstanding any copyright annotation therein.

\bibliographystyle{IEEEtran}
\footnotesize
\bibliography{IEEEabrv}

%

\begin{IEEEbiography}[{\includegraphics[width=1in,height=1.25in,clip,keepaspectratio]{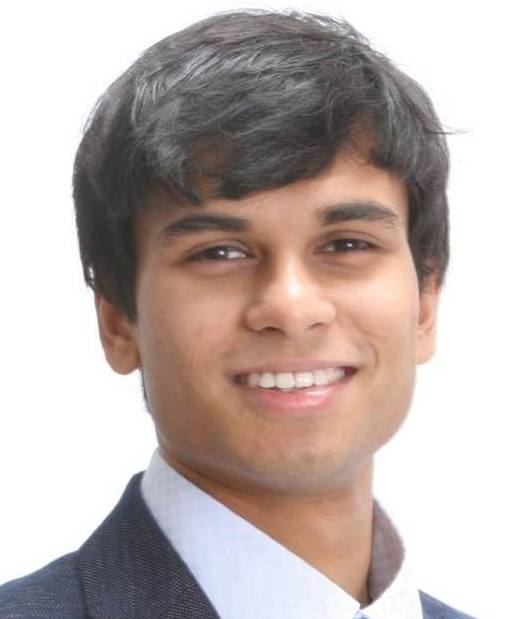}}]{Debayan Deb}
received his B.S. degree in  computer  science  from  Michigan State University,  East Lansing,  Michigan,  in  2016.  He  is currently working towards a PhD degree in the Department of Computer Science and Engineering at Michigan State University, East Lansing, Michigan. His research interests include pattern recognition, computer vision, and machine learning with applications in biometrics.
\end{IEEEbiography}

\begin{IEEEbiography}[{\includegraphics[width=1in,height=1.25in,clip,keepaspectratio]{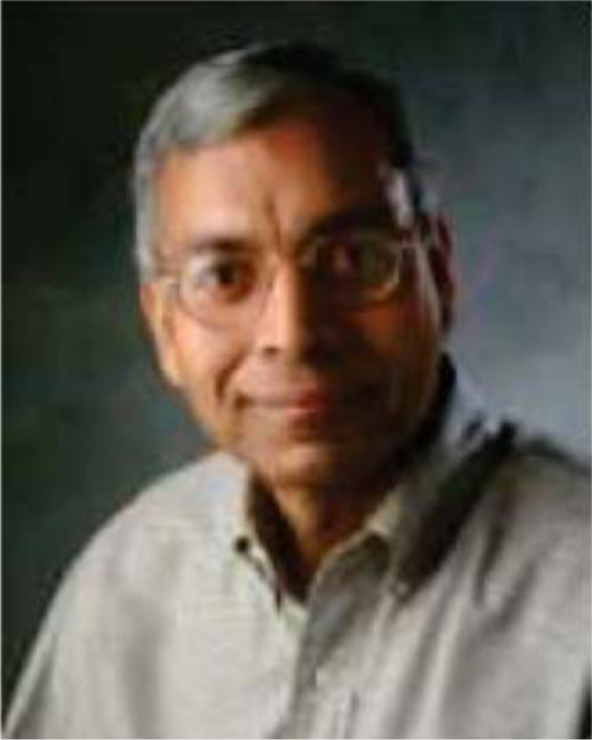}}]{Anil K. Jain}
is a University distinguished professor in the Department of Computer Science and Engineering at Michigan State University. His research interests include pattern recognition and biometric authentication. He served as the editor-in-chief of the IEEE Transactions on Pattern Analysis and Machine Intelligence and was a member of the United States Defense Science Board. He has received Fulbright, Guggenheim, Alexander von Humboldt, and IAPR King Sun Fu awards. He is a member of the National Academy of Engineering and foreign fellow of the Indian National Academy of Engineering and Chinese Academy of Sciences.
\end{IEEEbiography}





\end{document}